\documentclass[10pt,twocolumn,letterpaper]{article}
\usepackage[pagenumbers]{cvpr}   %

\usepackage{multirow}

\newcommand{\vf}[1]{{\color{orange}#1}}

\newcommand{\ours}{VFR\xspace}
\newcommand{\oursmi}{VFR+MI\xspace}       %
\newcommand{\base}{Base\xspace}
\newcommand{\uv}{Unverified\xspace}        %
\newcommand{\vbn}{VisMin-native\xspace}    %
\newcommand{\vbc}{VisMin-controlled\xspace}%
\newcommand{\qwen}{Qwen3-VL-8B-Instruct\xspace}

\DeclareMathOperator*{\argmax}{arg\,max}
\newcommand{\yes}{\texttt{Yes}\xspace}
\newcommand{\no}{\texttt{No}\xspace}

\renewcommand{\vf}[1]{#1}        %

\definecolor{cvprblue}{rgb}{0.21,0.49,0.74}
\usepackage[pagebackref,breaklinks,colorlinks,allcolors=cvprblue]{hyperref}
\hypersetup{
  pdftitle={When Repair Accuracy Misleads: Training a VLM on Its Own Failures},
  pdfauthor={Ang Qi, Yue Wang}
}

\def\confName{CVPR}
\def\confYear{2027}
\def\arxivversion{}              %

\newif\ifsupp
\supptrue

\title{When Repair Accuracy Misleads:\\Training a VLM on Its Own Failures}

\author{Ang Qi \qquad Yue Wang\\
USC Physical Superintelligence (PSI) Lab
}

\begin{document}
\maketitle

\begin{abstract}
Fine-tuning a vision-language model (VLM) on its own failures is a direct way to repair it, but accuracy on held-out failures, the obvious measure of success, misleads.
Many mined failures are not model errors: in a human audit of 100 failures that \qwen made on questions generated from COCO and GQA annotations, \vf{37} had a wrong stored label.
And mined failures share regularities in their wording: question-only classifiers get both answers right on \vf{73.2\%} of held-out failure pairs.
We study both traps with verified failure-driven repair (\ours), a deliberately simple pipeline: typed minimal-pair failure mining, verification by three LLM judges that answer each question without seeing its label, answer-first corrective targets rendered from the verified annotations, and LoRA training mixed with replay, mostly of the base model's own responses.
\ours repairs \vf{82.8\%} of image-disjoint held-out failures (base model: \vf{10.0\%}; five seeds), and on independently labeled, unconditioned pairs it raises overall accuracy from \vf{87.4\%} to \vf{92.4\%}.
The gain comes from spatial relations, whose accuracy rises from \vf{66.7\%} to \vf{96.8\%}, even where the question's wording points to the wrong answer.
Yet on negation, of which it repairs \vf{88.9\%} of held-out failures, its accuracy on unconditioned pairs falls from \vf{88.7\%} to \vf{79.7\%}: it breaks negations about absent objects, which are rare in its training data.
Without verification, accuracy on unconditioned pairs drops to \vf{80.8\%}, below the base model's \vf{87.4\%}, through a negation shortcut that our development benchmarks miss; without failure selection, it stays at \vf{92.0\%} (one seed) while repair falls to \vf{73.4\%}.
Training also lowers two-image relation accuracy, from \vf{74.3\%} to \vf{49.6\%} for \ours and to \vf{48.4\%} on the replay data alone; asking \ours to describe both images first recovers \vf{10.7} points on average across its five seeds, and multi-image replay partly and unevenly reverses the drop.
On four held-out benchmarks registered in advance, \ours gains \vf{2.9} points on CV-Bench 2D, loses \vf{3.3} on BLINK, mostly on its two-image tasks, and shows no clear change on CV-Bench 3D, MME, and MMVP.
Most of the spatial gain survives on the same spatial pairs with paraphrased questions (\vf{87.3\%} \vs\ the base model's \vf{62.0\%}), though rewording lowers spatial repair on held-out failures from \vf{92.9\%} to \vf{76.8\%}.
On an independent set of controlled photos (What'sUp), caption choice stays at the base model's level (\vf{97.7\%} \vs\ \vf{97.6\%}) while answers to our yes/no questions about objects under others collapse (from \vf{84.5\%} to \vf{26.4\%} of photos).
\end{abstract}

\section{Introduction}
\label{sec:intro}

\begin{figure*}[t]
  \centering
  \includegraphics[width=\linewidth]{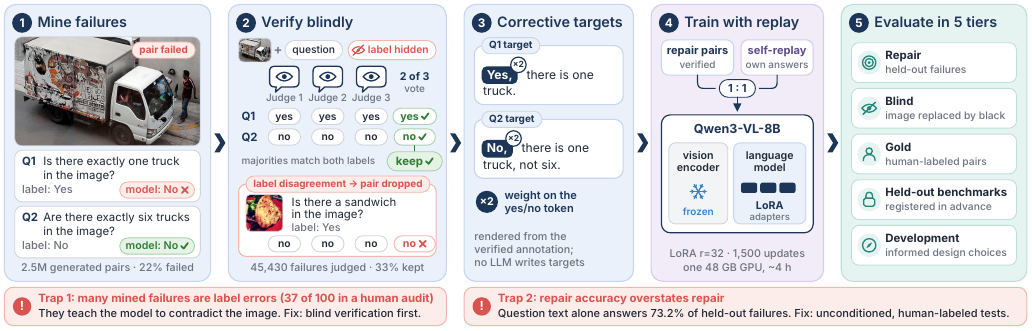}
  \caption{\textbf{Verified failure-driven repair.}
  (1)~We mine typed minimal pairs that the target VLM answers wrongly against the stored annotation; on this counting pair from the training split, it answers \no to Q1.
  (2)~Three LLM judges answer each question from the image without seeing its stored label; a pair is kept only if both two-of-three majorities agree with the stored labels.
  Inset: a rejected pair; its stored label puts a sandwich on this plate of chicken, and all three judges answer \no.
  (3)~Each kept question gets an answer-first, fact-grounded target rendered from the annotation, with extra weight on the yes/no token.
  (4)~We fine-tune with LoRA on repair pairs mixed 1:1 with self-replay, whose open-ended targets are the base model's own answers.
  (5)~We evaluate on five tiers.
  Orange boxes: the two traps and our response to each.
  Photos: COCO~\cite{coco} train2017 images 326700 and 402717 (Flickr photos 7574779034 and 8098728576; CC~BY~2.0), cropped.}
  \label{fig:pipeline}
\end{figure*}

Vision-language models (VLMs) answer open-ended questions about images fluently, yet still fail simple, typed questions: whether an object is present, what color it is, which side of another object it is on, how many there are, or whether a negated statement holds~\cite{pope,repope,hallusionbench,naturalbench,mmvp,vismin,negvqa}.
For a given model these failures are systematic.
We generated 2.5 million minimal pairs (two yes/no questions about one image that differ in a single factor) from COCO and GQA annotations~\cite{coco,gqa}; \qwen~\cite{qwen3vl} was scored wrong on at least one question of \vf{22.2\%} of the pairs, and of \vf{53.7\%} of the spatial-relation pairs.

Most remedies are model-agnostic.
Robust instruction data~\cite{lrv}, preference training~\cite{rlhfv,mdpo}, and contrastive decoding~\cite{vcd,opera} target hallucination in general, and minimal-change training sets such as VisMin~\cite{vismin} are built without reference to what a particular model gets wrong.
Recent work instead builds training data from a model's own errors, by synthesizing examples that target its reasoning failures~\cite{reasoningfailures} or by turning its hallucinated responses into preference pairs~\cite{ovip}.
The simplest version of this idea is to mine a model's failures on typed questions and train on them directly.
We find that it can repair spatial relations but has two traps.
First, many mined failures are not model errors.
In a human audit of 100 mined failures, only \vf{45} were clear model errors; \vf{37} had a stored label that the image contradicts, and \vf{18} were ambiguous.
Training on such pairs teaches the model to contradict the image.
Second, the natural measure of success, \emph{repair accuracy} (the share of held-out failures answered correctly after training), overstates what training repairs.
Mined failures share regularities in their wording, so much of repair accuracy can be earned without the image: classifiers that see only the question text, trained on our training failures, get both answers right on \vf{73.2\%} of held-out failure pairs.
And held-out failures rarely contain what training breaks: our verified model repairs \vf{88.9\%} of held-out negation failures, yet on unconditioned, human-labeled pairs its negation accuracy falls from \vf{88.7\%} to \vf{79.7\%}.

We study both traps with verified failure-driven repair (\ours, \cref{fig:pipeline}), a deliberately simple pipeline.
We mine typed minimal pairs on which the target model fails, and three LLM judges answer the two questions of a mined pair from the image without seeing the stored labels; the pair is kept only if their majority answers agree with them.
Each verified question receives an answer-first corrective target rendered deterministically from the verified annotation (\eg, ``No, there is one truck, not six.''), and we fine-tune with LoRA~\cite{lora} on these pairs mixed one-to-one with general instructions, whose open-ended answers are the base model's own.
Because repair accuracy alone misleads, we evaluate on five tiers: repair on held-out failures next to a question-only classifier, a blind diagnostic with a black image, a gold test of unconditioned pairs labeled blind by a human and split by whether the wording alone points to the right answer, four held-out benchmarks registered in advance, and five development benchmarks.

Our contributions are empirical findings about this setting:
\begin{itemize}
  \item \textbf{A verified repair recipe that raises accuracy on independently labeled pairs.}
  \ours repairs \vf{82.8\%} of image-disjoint held-out failures of \qwen (base model \vf{10.0\%}; TEST: \vf{82.9\%}), against \vf{27.0\%} for VisMin minimal-change data with the same budget and recipe.
  On gold it gains \vf{5.0} points (95\% interval \vf{2.5}--\vf{7.5}), a gain that comes from spatial relations (\vf{66.7}$\to$\vf{96.8\%}), which improve even where the wording points to the wrong answer.
  On held-out benchmarks registered in advance, it gains \vf{2.9} points on CV-Bench 2D and loses \vf{3.3} on BLINK.
  \item \textbf{Mined failures need verification.}
  Training on unverified mined failures repairs \vf{6.8} points less and teaches a negation shortcut that none of our development benchmarks detects: on unconditioned gold pairs, the unverified model falls below the base model (\vf{80.8\%} \vs\ \vf{87.4\%}; \ours: \vf{92.4\%}) and answers \vf{4.0\%} of negations about absent objects correctly (base model: \vf{88.0\%}).
  \item \textbf{Repair accuracy overstates repair.}
  Where the wording points to the wrong answer, gold spatial accuracy rises from \vf{70.8\%} to \vf{97.5\%}, but pooled over all five types \ours shows no clear gain over the base model, and its negation repair on held-out failures hides a loss on gold, concentrated in negations about absent objects (\vf{88.0}$\to$\vf{65.9\%}), which are rare in its training data.
  Selecting failures adds \vf{8.5} points of TEST repair over verified pairs drawn regardless of failure, but only \vf{0.8} on gold (interval \vf{$-0.8$} to \vf{2.5}; seed 42).
  Recipes with equal repair also differ in their answer prior and in how much they answer from the question alone.
  \item \textbf{Fine-tuning degrades an untrained task.}
  Two-image relation accuracy on VisMin falls from \vf{74.3\%} to \vf{49.6\%} (five seeds), and to \vf{48.4\%} with the replay data alone, as the model picks the same image for opposite captions; over five seeds, describing both images first recovers \vf{10.7} points on average (\vf{6.1} to \vf{15.4}) and multi-image replay \vf{12.4} (\vf{2.6} to \vf{28.9}) at essentially no repair cost.
  On held-out BLINK, its two-image tasks fall by \vf{10.8} points on average.
  \item \textbf{The spatial gain is largest on above/below and carries over in part.}
  On gold, above/below accuracy rises from \vf{42.9\%} to \vf{95.6\%}, while left/right was already at \vf{91.8\%}; with paraphrased gold questions \ours keeps most of its gain across both axes (\vf{87.3\%} \vs\ the base model's \vf{62.0\%}), while rewording lowers spatial repair on held-out failures from \vf{92.9\%} to \vf{76.8\%}, and CV-Bench 2D's gain is partly in counting.
  On What'sUp's controlled photos, caption choice stays at the base model's level (\vf{97.7\%} \vs\ \vf{97.6\%}), but every arm trained on our pairs answers our yes/no questions about objects under others worse (\ours: from the base model's \vf{84.5\%} to \vf{26.4\%} of photos).
\end{itemize}
Code, data, prompts, and the registered protocols will be released.

\section{Related work}
\label{sec:related}

\paragraph{Typed failures and hallucination in VLMs.}
Object hallucination is measured with caption metrics such as CHAIR~\cite{chair} and with yes/no probes such as POPE~\cite{pope}, whose label errors RePOPE corrects~\cite{repope}; AMBER~\cite{amber} also covers attributes and relations.
Paired and adversarial benchmarks expose failures that aggregate scores hide: HallusionBench~\cite{hallusionbench}, NaturalBench~\cite{naturalbench}, and MMVP~\cite{mmvp} for fine-grained visual questions; Winoground~\cite{winoground}, SugarCrepe~\cite{sugarcrepe}, and What'sUp~\cite{whatsup} for composition and spatial relations; NegBench~\cite{negbench} and NegVQA~\cite{negvqa} for negation; and VisMin~\cite{vismin} for minimal visual changes.
Models can also answer many such questions from language priors alone~\cite{agrawal2018priors,mmstar}.
These works diagnose failures; we repair one model's typed failures and use such benchmarks only for evaluation.

\paragraph{Training against hallucination.}
LRV-Instruction~\cite{lrv} adds negative and manipulated instructions to instruction tuning; RLHF-V~\cite{rlhfv} and mDPO~\cite{mdpo} train on preferences; VCD~\cite{vcd} and OPERA~\cite{opera} change decoding without training.
Closer to our setting, some methods build training data from the target model's own errors: Stan \etal~\cite{reasoningfailures} have frontier models analyze a weaker multimodal model's reasoning failures and generate filtered corrective examples, and OViP~\cite{ovip} builds preference pairs online from the model's own hallucinated responses.
We study the simplest version, training directly on a model's mined failures on typed questions, and analyze what it teaches.

\paragraph{Minimal pairs, failure discovery, and label noise.}
Contrast sets~\cite{gardner2020contrast} and counterfactually augmented data~\cite{kaushik2020cad} use minimal edits to probe and train text models.
VisMin~\cite{vismin} releases minimal-change image--caption data and fine-tunes CLIP and Idefics2~\cite{idefics2} on it; it is our main data baseline.
Slice discovery~\cite{domino} and model patching~\cite{goel2021patching} find and repair a given model's systematic errors.
Label errors are common even in test sets~\cite{northcutt2021pervasive}, including the COCO-derived POPE~\cite{repope}; mining a model's failures selects for them, because a correct answer scored against a wrong label counts as a failure, and training then propagates them.
This motivates our verification step.
LLMs are widely used as judges~\cite{zheng2023judging}, and panels of different judges reduce single-model bias~\cite{verga2024juries}.
Our committee is such a panel, but it answers each question from the image instead of grading a response.

\paragraph{Forgetting and replay.}
Fine-tuning multimodal LLMs can cause catastrophic forgetting~\cite{zhai2024forgetting}; replaying earlier data~\cite{rolnick2019replay} and training on the model's own outputs~\cite{yang2024sdft} limit it.
We replay general instructions with the base model's own responses as targets.
The side effect we report, a bare choice that often picks the same image for opposite captions, resembles shortcut learning~\cite{geirhos2020shortcut}.

\section{Verified failure-driven repair}
\label{sec:method}

\subsection{Typed minimal pairs}
\label{sec:setup-pairs}
A \emph{typed minimal pair} $(I,q_1,q_2,a_1,a_2)$ is an image $I$ with two yes/no questions that differ in a single factor and have opposite answers.
Each image--question pair is a \emph{cell}, and a pair counts as correct only if both of its cells are.
We use five types (examples in Supp.~\ref{supp:mining}).
\emph{Object existence} contrasts an annotated object with an absent one.
\emph{Attribute swap} contrasts an object's color with a different color (``Is there a white drape in the image?'' / ``Is there a purple drape in the image?'').
\emph{Spatial relation} swaps left/right or above/below between two annotated objects in the image frame (``In the image, is there a car above a skateboard?'' / ``\dots below a skateboard?'').
\emph{Counting} contrasts the number of objects stated in a caption and matched by the instance annotations (1--9 in TRAIN) with a different number at most five away from it, as in ``Is there exactly one toilet in the image?'' / ``Are there exactly six toilets in the image?''.
\emph{Negation} negates an existence, attribute, or relation fact about a present or an absent object (six operators; \eg, ``In the image, is there no toaster to the left of a knife?'').
Pairs are generated from COCO train2017 instances and captions~\cite{coco} and from GQA scene graphs~\cite{gqa}.
Each pair stores its changed factor (\eg, the true and the asked color), from which its training targets are later rendered.
A model answers a cell with $\argmax_{y\in\{\yes,\no\}}\log P(y\mid I,q)$ at the first answer position, under a fixed instruction wrapper; ties, where the two log probabilities are within $\smash{10^{-8}}$ of each other, count as wrong.

\subsection{Mining the model's failures}
\label{sec:mining}
We score \qwen on the generated pairs with a one-word wrapper (``Answer the question using only one word: Yes or No.''; all wrappers in Supp.~\ref{supp:mining}).
A pair is a \emph{mined failure} if at least one of its cells disagrees with the stored label; there is no margin threshold, and pairs with a tie are excluded.
We mined in two rounds: \vf{69{,}000} pairs, of which \vf{25.3\%} failed, and \vf{2.5}M new pairs (\vf{500}k per type), of which \vf{22.2\%} failed.
Failure rates differ strongly by type, from \vf{6.7\%} for counting to \vf{53.7\%} for spatial relations in round two.
The verified pool has \vf{15{,}000} pairs (\vf{3{,}000} per type), which we split into image-disjoint TRAIN, VAL, and TEST sets of \vf{12{,}000}, \vf{1{,}500}, and \vf{1{,}500} pairs on \vf{9{,}528}, \vf{1{,}336}, and \vf{1{,}330} distinct images, each balanced by type.

\subsection{Committee verification}
\label{sec:verification}
A mined failure is either a model error or a label problem, where the stored label disagrees with the image (\eg, a missed annotation or a wrong color).
Training on label problems teaches the model to contradict the image.
We therefore verify mined failures with three LLM judges: Claude Sonnet~5~\cite{claudesonnet5}, Seed2.1~Pro~\cite{seed21}, and Gemini~3.6~Flash~\cite{gemini36flash}.
Each judge answers each question separately from the image, with \emph{yes}, \emph{no}, or \emph{uncertain}, and sees neither the stored label nor the model's answer (prompt in Supp.~\ref{supp:committee}).
A two-of-three majority decides each question.
A pair is kept only if both majorities agree with the stored labels.
It is dropped as a \emph{label disagreement} if a majority contradicts a stored label, and as \emph{ambiguous} if a question has no majority or an \emph{uncertain} majority.
The committee is a filter, not ground truth: a label disagreement shows that the judges contradict the label, not that the model was right.

We sent the committee all round-one failures and a random sample of round-two ones, drawn per type until each type had at least 3{,}000 kept pairs.
Of these \vf{45{,}430} failures, it kept \vf{33.1\%}, flagged \vf{57.8\%} as label disagreements, and found \vf{9.0\%} ambiguous.
Keep rates range from \vf{25.2\%} (object existence) to \vf{39.5\%} (negation).
The committee keeps \vf{72.5\%} of pairs drawn regardless of failure: \vf{45.6\%} of the model's failures and \vf{88.7\%} of its successes, so mining concentrates the pairs it rejects.
Two human audits by one of the authors check the committee from both sides.
Before verification, 100 mined failures (20 per type) contained \vf{45} model errors, \vf{37} wrong labels (Wilson 95\% CI [\vf{28.2}, \vf{46.8}]), and \vf{18} ambiguous pairs; an earlier version of the committee, with Gemini~3.5~Flash as the third judge, kept these model errors with \vf{80\%} precision and \vf{71\%} recall, and the final committee keeps them with \vf{85.7\%} precision and \vf{66.7\%} recall and rejects \vf{36} of the \vf{37} wrong labels (Supp.~\ref{supp:committee}).
These pairs were the committee's calibration set, so this is calibration evidence, not an independent test.
The base model's own confidence cannot stand in for the committee: its AUROC for wrong labels is \vf{0.76} on the audited pairs and \vf{0.72} against the committee (Supp.~\ref{supp:committee}).
After verification and target rendering, 100 training pairs had \vf{88.0\%} correct target cells [\vf{83.0}, \vf{93.0}] and \vf{5.0\%} factually wrong ones [\vf{1.5}, \vf{9.0}].

\subsection{Corrective fact-grounded targets}
\label{sec:targets}
The simplest training target for a yes/no cell is the bare answer.
We instead train with \emph{corrective} targets that give the answer first and then the verified fact, rendered deterministically from the stored factor with eight templates per type (19 for negation); no LLM writes targets.
The counting pair above yields ``Yes, the image shows one toilet.'' and ``No, there is one toilet, not six.''
Training uses a wrapper that asks for the answer and a one-sentence explanation (Supp.~\ref{supp:mining}).
Every target begins with the same first token as the bare answer, so first-token scoring is unchanged (Supp.~\ref{supp:targets}).

A corrective target dilutes the decision.
Let $\ell_t$ be the cross-entropy of target token $t$ in a target of length $L$ whose first token is the \yes/\no decision.
We weight the decision token by $w$:
\begin{equation}
  \begin{aligned}
  \mathcal{L}_{\text{cell}} &= \frac{w\,\ell_1 + \sum_{t=2}^{L} \ell_t}{w + L - 1},\\
  \mathcal{L}_{\text{pair}} &= \tfrac{1}{2}\big(\mathcal{L}_{\text{cell}}(q_1) + \mathcal{L}_{\text{cell}}(q_2)\big).
  \end{aligned}
  \label{eq:cellloss}
\end{equation}
The decision token's share of the weights in a cell's loss is $w/(w+L-1)$.
At the mean target length of \vf{10.87} tokens, this share is \vf{9.2\%} for $w{=}1$ and \vf{16.9\%} for $w{=}2$, the value we use (\cref{sec:selection}).

\subsection{Training with self-replay}
\label{sec:replay}
We fine-tune with LoRA~\cite{lora} on an equal mix of repair pairs and replay rows,
$\mathcal{L}=\tfrac12\,\mathbb{E}_{p}[\mathcal{L}_{\text{pair}}]+\tfrac12\,\mathbb{E}_{r}[\mathcal{L}_{\text{replay}}]$,
where a replay row's loss is its token mean and the end-of-turn token is not supervised in either part.
The 12{,}000 replay rows come from eight categories (Supp.~\ref{supp:replay}): descriptions of LLaVA and COCO images (\vf{33\%}); open-ended VQA from LLaVA~\cite{llava} and non-binary VQAv2~\cite{vqav2} (\vf{26\%}); balanced VQAv2 yes/no questions (\vf{24\%}); Dolly text-only instructions~\cite{dolly} (\vf{8\%}); ChartQA~\cite{chartqa} (\vf{5\%}); ScienceQA~\cite{scienceqa} (\vf{3\%}); and small negation and counting yes/no sets (\vf{0.6\%}).
Descriptions, open-ended VQA, and text-only instructions (\vf{8{,}020} rows) use the frozen base model's own greedy responses as targets, so replay ties the model to its own behavior rather than to another model's style; the other categories keep their dataset answers.
LoRA (rank 32, $\alpha{=}64$, dropout 0.05) adapts the attention and MLP projections of all 36 language-model layers (\vf{87.3}M trainable parameters, \vf{1.0\%} of the model); the vision encoder and projector stay frozen.
We train for one epoch, \vf{1{,}500} updates of 8 repair pairs and 8 replay rows, with AdamW, learning rate $5{\times}10^{-5}$, 45 warm-up updates, cosine decay, and gradient clipping at 1.0; training takes \vf{4.1} hours on one 48\,GB GPU.

\section{Experimental setup}
\label{sec:exp}

\begin{table*}[t]
  \caption{\textbf{Main results.}
  \emph{Repair}: pair accuracy on held-out failures of the base model (frozen VAL; a repair rate, not accuracy).
  \emph{Gap}: repair minus pair accuracy with the image replaced by a black one.
  \emph{Gold}: unconditioned pairs with blind human labels; \emph{Conf.}: the \vf{244} gold pairs whose wording points to the wrong answer (the text-only classifiers get them wrong); \emph{Regr.}: share of base-correct gold pairs answered wrongly.
  Held-out benchmarks (robust tier) were registered before any adapter was evaluated and informed no decision; development benchmarks did, with their official scorers (for MMBench, see \cref{sec:main}).
  \emph{Text only}: logistic regressions on the question text alone, fit post hoc on TRAIN; \oursmi: \ours with multi-image replay (\cref{sec:sideeffect}); --: not reported or not applicable.
  Internal and gold columns: mean over five seeds for \ours and \oursmi and three for the VisMin arms and \uv; replay only and random pairs (\cref{sec:exp}): seed 42; external columns: seed 42.}
  \label{tab:main}
  \centering
  \footnotesize
  \setlength{\tabcolsep}{3.2pt}
  \begin{tabular}{@{}l cc ccc cccc cccc@{}}
    \toprule
    & \multicolumn{2}{c}{Internal} & \multicolumn{3}{c}{Gold} & \multicolumn{4}{c}{Held-out} & \multicolumn{4}{c}{Development} \\
    \cmidrule(lr){2-3}\cmidrule(lr){4-6}\cmidrule(lr){7-10}\cmidrule(lr){11-14}
    Method & Repair$\uparrow$ & Gap & Acc$\uparrow$ & Conf.$\uparrow$ & Regr.$\downarrow$
      & MME P/C$\uparrow$ & CV-B 2D/3D$\uparrow$ & MMVP$\uparrow$ & BLINK$\uparrow$
      & MMB$\uparrow$ & NB-G$\uparrow$ & HB-q$\uparrow$ & RePOPE$\uparrow$ \\
    \midrule
    \base  & \vf{10.0} & \vf{8.1} & \vf{87.4} & \vf{84.8} & -- & \vf{1739/657} & \vf{78.4/93.6} & \vf{59.3} & \vf{68.5}
           & \vf{77.6} & \vf{37.0} & \vf{42.9} & \vf{93.2} \\
    Text only & \vf{73.2} & -- & \vf{67.3} & -- & -- & -- & -- & -- & -- & -- & -- & -- & -- \\
    \uv    & \vf{76.0} & \vf{27.2} & \vf{80.8} & \vf{57.4} & \vf{17.6} & \vf{1748/661} & \vf{80.7/93.1} & \vf{61.3} & \vf{66.4}
           & \vf{83.6} & \vf{39.3} & \vf{42.0} & \vf{93.2} \\
    \vbn   & \vf{17.5} & \vf{15.8} & \vf{86.9} & \vf{84.0} & \vf{2.6} & \vf{1731/669} & \vf{80.1/93.5} & \vf{61.3} & \vf{63.3}
           & \vf{77.8} & \vf{40.4} & \vf{42.4} & \vf{92.9} \\
    \vbc   & \vf{27.0} & \vf{18.3} & \vf{84.5} & \vf{81.7} & \vf{6.2} & \vf{1739/676} & \vf{80.5/93.4} & \vf{62.0} & \vf{66.5}
           & \vf{84.1} & \vf{40.2} & \vf{43.7} & \vf{92.9} \\
    Replay only & \vf{9.9} & \vf{8.4} & \vf{85.3} & \vf{81.6} & \vf{3.5} & \vf{1726/653} & \vf{80.3/93.0} & \vf{61.3} & \vf{67.0}
           & \vf{83.5} & \vf{39.3} & \vf{43.5} & \vf{93.1} \\
    Random pairs & \vf{73.4} & \vf{40.7} & \vf{92.0} & \vf{82.0} & \vf{4.3} & \vf{1766/661} & \vf{80.9/93.5} & \vf{61.3} & \vf{65.8}
           & \vf{84.3} & \vf{39.7} & \vf{42.9} & \vf{93.4} \\
    \midrule
    \ours  & \vf{82.8} & \vf{35.3} & \vf{92.4} & \vf{82.4} & \vf{4.9} & \vf{1757/661} & \vf{81.4/93.6} & \vf{61.3} & \vf{65.2}
           & \vf{84.1} & \vf{39.9} & \vf{41.5} & \vf{93.0} \\
    \oursmi & \vf{82.8} & \vf{35.3} & \vf{92.0} & \vf{80.9} & \vf{5.2} & \vf{1757/661} & \vf{81.1/93.3} & \vf{60.7} & \vf{66.3}
           & \vf{83.5} & \vf{39.1} & \vf{42.0} & \vf{92.8} \\
    \bottomrule
  \end{tabular}
\end{table*}

\paragraph{Model.}
We repair \qwen~\cite{qwen3vl} at a pinned revision in bf16 (Supp.~\ref{supp:eval}).
All arms share its tokenizer, processor, and image-resolution policy.

\paragraph{Evaluation tiers.}
No single number measures repair, so we report five tiers, each answering a different question.
\emph{(i) Repair:} pair accuracy on the \vf{1{,}500} VAL pairs, all of which the base model failed at mining.
This is a repair rate, not an estimate of accuracy, and VAL also informed our recipe decisions (\cref{sec:selection}).
Evaluation uses a slightly different wrapper from mining (``Answer the question based only on what is visible in the image. Answer Yes or No.''), under which the base model passes \vf{10.0\%} of VAL pairs; at mining, their \vf{153} wrong cells had favored the wrong answer by a median log-odds of \vf{0.5} (Supp.~\ref{supp:mining}).
Rescored with the mining wrapper, under which the base model fails every VAL pair, \ours (seed 42) repairs \vf{81.2\%} (evaluation wrapper: \vf{82.8\%}).
A \emph{rephrased VAL} asks the same pairs in new template wording; it was registered before any VLM was evaluated on it (Supp.~\ref{supp:eval}).
\emph{(ii) Blind diagnostic:} the same pairs with each image replaced by a black image of the same size, passed through the normal processor.
Blind accuracy measures how much the model can answer from the question alone, and the \emph{sighted--blind gap} is sighted minus blind pair accuracy.
A black image is itself an unusual input, so neither number directly measures grounding.
\emph{(iii) Gold test:} \vf{747} pairs (\vf{150} per type; \vf{147} for object existence) sampled from all generated pairs, whether or not the model fails them, on images disjoint from all training and evaluation data.
One of the authors labeled them from the image alone, blind to the generator's labels and to every model output, without adjudication.
We dropped pairs with a ``can't tell'' answer, an inconsistent answer to a repeated item, or the same answer to both questions (counts in Supp.~\ref{supp:eval}), so the set is exactly half \yes.
For every arm, including all variants in \cref{tab:ablations}, we report pair accuracy and regressions (pairs the base model gets right that the trained model gets wrong), also on the \vf{244} pairs that the question-only classifiers of \cref{sec:main} answer wrongly, whose wording points to the wrong answer.
\emph{(iv) Held-out benchmarks:} MME~\cite{mme}, CV-Bench~\cite{cambrian}, MMVP~\cite{mmvp}, and the BLINK validation set~\cite{blink}.
We registered them, with their scorers, exclusions, and gates, before any adapter was evaluated on them, and they informed no decision.
After decontamination they have \vf{2{,}310}, \vf{1{,}260}\,+\,\vf{1{,}200} (2D\,+\,3D), \vf{300}, and \vf{1{,}772} questions.
To test whether the spatial repair carries over to new photos, we added What'sUp's two controlled subsets~\cite{whatsup}, which photograph each pair of objects in four relations: on, under, left of, and right of each other in one subset, and in front of, behind, left of, and right of in the other; we specified this test after the main results and registered it before running it (Supp.~\ref{supp:eval}).
We pose its photos as four-way caption choice, in all four cyclic orders of the options because the release lists the correct caption first, and in our yes/no template, asking each photo about the true relation and its opposite (\eg, under \vs\ on) and counting the photo as correct only if both answers are, with intervals over object pairs.
\emph{(v) Development benchmarks:} MMBench-Dev-EN v1.1 with circular evaluation~\cite{mmbench}, NaturalBench group accuracy~\cite{naturalbench}, HallusionBench question accuracy~\cite{hallusionbench}, RePOPE macro F1, yes-rate, and precision~\cite{repope}, and AMBER~\cite{amber}.
These informed our design and hyperparameter decisions.
We also probe NegVQA~\cite{negvqa} and VisMin-bench~\cite{vismin}; the latter is in-distribution for the VisMin-trained baselines.
TEST was used only before training, for release validation, a data audit, and a pre-registered base-model diagnostic, and to exclude its images from all later data.
After freezing the recipe, we evaluate the final arms and all variants in \cref{tab:ablations} on TEST once; no decision follows from it.

\paragraph{Answer extraction and decontamination.}
Benchmarks use their native prompts, greedy decoding, and pinned harness and scorers~\cite{vlmevalkit}.
Strict official parsers can mark a correct full-sentence answer wrong, so for held-out benchmarks we report both the official scorer and a \emph{robust} tier.
In the robust tier, a response that is exactly one option label is scored directly; any other response goes to a locked local LLM extractor, Muse Glimmer~30B~\cite{muse}, which agreed with a human on \vf{79} of \vf{80} capability-probe responses and on \vf{59} of \vf{60} free-form held-out responses, half of which the official parsers could not resolve.
For each checkpoint, up to 200 randomly chosen exact-label responses also go to the extractor as canaries, and unresolved responses count as wrong.
We remove benchmark items whose images appear in any training or selection data (failure pairs, replay data, VisMin edit families, validation sets) or in other evaluations, matching by source identity, SHA-256, or perceptual hash.

\paragraph{Baselines.}
The three data baselines use \ours's base model, LoRA recipe, budget (12{,}000 repair units, the same 12{,}000 replay rows, 1{,}500 updates), and seeds (42--44) unless stated otherwise.
\emph{\uv} applies our recipe to 12{,}000 mined failures (2{,}400 per type) drawn from the same failure pool without the committee.
\emph{\vbn} trains on VisMin's released minimal-change data~\cite{vismin}, reconstructed with the authors' builder as 6{,}000 two-image and 6{,}000 two-caption units of two choice questions each, with a uniform token loss; we omit its captioning rows.
\emph{\vbc} converts 12{,}000 VisMin edits (3{,}000 each of object, attribute, counting, and relation) to our single-image yes/no format (``Does the image match this description: \dots?'') with corrective targets and decision weighting, which isolates the data source from the format.
VisMin data is only decontaminated (\vf{41{,}117} of \vf{64{,}392} edits remain) and converted, never filtered with our committee.
Two seed-42 controls train on the 12{,}000 replay rows alone (\emph{replay only}) or replace the mined failures with \vf{12{,}000} training pairs (\vf{2{,}400} per type) drawn at random regardless of failure (\vf{22.9\%} are base-model failures) and kept by the same committee (\emph{random pairs}; \cref{sec:verification}).

\paragraph{Statistics.}
For internal and gold metrics we report means and ranges over seeds (five for \ours and \oursmi, three for the VisMin arms and \uv), and for comparisons between arms paired bootstrap intervals over image clusters (\vf{20{,}000} draws for the VAL contrast of \ours with \uv, \vf{2{,}000} for the others), computed on seed means; they are conditional on the trained checkpoints and not adjusted for multiple comparisons.
The multi-image replication instead uses a $t$ interval over five matched seeds (\cref{sec:sideeffect}).
Differences are computed before rounding.
For benchmark comparisons we use a paired bootstrap over image clusters (object pairs on What'sUp; 2{,}000 replicates)~\cite{efron1993bootstrap}, and \ours's seed spread calibrates single-seed differences (three seeds, official tier: \vf{0.7}--\vf{1.6} points on CV-Bench, MMVP, and BLINK).

\section{Results}
\label{sec:results}

\subsection{Repair on held-out failures}
\label{sec:main}
\cref{tab:main} compares \ours with the base model and three baselines.
\ours repairs \vf{82.8\%} of held-out failures (five seeds, \vf{82.1}--\vf{83.5\%}), up from \vf{10.0\%}, and every type improves: spatial relation \vf{93.0\%}, counting \vf{90.0\%}, negation \vf{88.9\%}, object existence \vf{76.3\%}, and attribute \vf{66.1\%} (five-seed means).
On TEST, evaluated once after the recipe was frozen, it repairs \vf{82.9\%} (three seeds; base model \vf{10.9\%}).
With the image replaced by a black one, \ours still answers \vf{47.5\%} of VAL pairs correctly (base model: \vf{1.9\%}), so training also teaches regularities of the question templates; sighted accuracy is \vf{35.3} points higher (the sighted--blind gap; base model: \vf{8.1}).
With another VAL image instead, \ours (seed 42) answers \vf{46.2\%} correctly (black image: \vf{48.3\%}; on negation, \vf{45.3\%} \vs\ \vf{70.0\%}).
The gap is largest for spatial relations (\vf{64.2}) and smallest for counting (\vf{13.4}), where \vf{76.6\%} of pairs are answered correctly without the image: only a count of one is asked in the singular, and answering \yes to singular questions and \no to plural ones solves \vf{78.7\%} of VAL counting pairs and \vf{91.3\%} of gold ones, so this cue comes from our generator.
More broadly, logistic regressions on the question text alone, fit post hoc on TRAIN, answer \vf{73.2\%} of VAL pairs correctly (negation: \vf{93.7\%}) but only \vf{67.3\%} of gold pairs (\cref{tab:gold}), where negation falls to chance, so the negation cue is specific to the mined failures.
Repair on mined failures thus partly measures how well a model learns these cues; \ours exceeds the text-only baseline on VAL by \vf{9.6} points.
Rephrasing the VAL questions while keeping their images and labels lowers \ours's repair from \vf{82.8\%} to \vf{75.4\%} (seeds 42--44; terse targets lose only \vf{1.8} points, \cref{sec:selection}), most on spatial relations (\vf{92.9}$\to$\vf{76.8\%}), so part of it is tied to our templates' wording.
The base model, whose failures on the original wording define VAL, passes \vf{25.1\%} of the rephrased pairs (\vf{57.7\%} of counting pairs, whose false number also changes; Supp.~\ref{supp:eval}).

\emph{Data source.}
VisMin minimal-change data, trained with the same recipe and budget, repairs \vf{27.0\%} in our format and \vf{17.5\%} in its native choice format, most where its edits overlap our types (counting \vf{53.8\%}, spatial relations \vf{35.4\%}), and lowers negation, which VisMin does not cover, from \vf{15.3\%} to \vf{8.0\%}.
This comparison is on our failure distribution and templates, the home ground of failure-driven data, and VisMin also differs in images, wording, and coverage, so it shows only that generic minimal-change data does not repair these failures.
Random pairs test failure selection: they repair \vf{73.4\%} of VAL and \vf{74.8\%} of TEST (\ours, seed 42: \vf{82.8\%} and \vf{83.3\%}; TEST difference \vf{$-8.5$}, interval \vf{$-10.5$} to \vf{$-6.6$}), less mainly on negation and object existence.
Few of their training negations concern spatial relations or attributes (\vf{3.2\%}, against \vf{45.7\%} of VAL's; only our first generation round produced these), yet most of the negation gap (\vf{38} of \vf{54} TEST pairs) is on negated existence of present objects, which they trained on more often than \ours; there, as on object existence, \ours answers far more TEST pairs correctly without the image (\vf{63.9\%} \vs\ \vf{7.6\%}; \vf{35.3\%} \vs\ \vf{3.3\%}).

\emph{Retention.}
On development benchmarks, \ours gains \vf{6.5} points on MMBench with its official parser and \vf{2.9} on NaturalBench, and changes HallusionBench and RePOPE F1 by \vf{$-1.3$} and \vf{$-0.2$} (seed 42).
MMBench rises about 6 points for every trained arm except \vbn (\vf{+0.2}), but its official parser fails on \vf{7.4\%} of the base model's responses (\ours: \vf{1.2\%}); on the \vf{1{,}132} groups it parses for both, each answers \vf{1{,}000} correctly.
With the robust tier's extractor (\cref{sec:exp}), the base model scores \vf{85.6\%} and \ours\ \vf{85.3\%} (difference \vf{$-0.3$}, interval \vf{$-1.4$} to \vf{$+0.8$}; over seeds 42--44: \vf{85.2\%}, \vf{$-0.4$}, \vf{$-1.4$} to \vf{$+0.7$}), so \ours's official gain comes from the parser.
The same holds for replay alone (robust: \vf{85.4\%}; difference \vf{$-0.2$}, interval \vf{$-1.2$} to \vf{$+0.8$}), training without replay (\vf{85.0\%}; \vf{$-0.6$}, \vf{$-2.0$} to \vf{$+0.8$}), and \vbn over seeds 42--44 (\vf{85.2\%}; \vf{$-0.4$}, \vf{$-1.5$} to \vf{$+0.6$}, against an official gain of \vf{3.0} points); the other arms were scored with the official parser only.
Without replay (\cref{tab:ablations}), \ours loses \vf{13.8} points on NaturalBench, \vf{8.4} on HallusionBench, and \vf{14.6} in RePOPE F1, so replay largely prevents these losses; replay alone repairs \vf{9.9\%} of VAL (base model: \vf{10.0\%}).
On the held-out benchmarks too, training without replay loses \vf{15.3} points on MMVP and \vf{3.7} on BLINK relative to \ours (robust tier, seed 42; intervals \vf{$-22.5$} to \vf{$-8.0$} and \vf{$-5.5$} to \vf{$-2.0$}), while it gains \vf{1.9} on CV-Bench 2D (an interval above zero).
Because the development benchmarks informed our decisions (\cref{sec:selection}), the held-out benchmarks carry the retention claim.
There (robust tier, seed 42), \ours gains \vf{2.9} points on CV-Bench 2D (95\% interval \vf{1.1}--\vf{4.8}), shows no clear change on CV-Bench 3D, MME, and MMVP, and loses \vf{3.3} on BLINK (\vf{1.4}--\vf{5.1}), most on its two-image tasks (\cref{sec:sideeffect}).
Both changes hold over seeds 42--44 (CV-Bench 2D: \vf{+2.9}, interval \vf{1.2}--\vf{4.6}; BLINK: \vf{$-2.8$}, \vf{$-4.6$} to \vf{$-1.2$}), and in each seed CV-Bench 3D, MME, and MMVP show no clear change.
Official parsers, which fail on more of the base model's long answers (BLINK: \vf{12.7\%} unresolved; \ours: none), would favor \ours on both (CV-Bench 2D: \vf{+6.8}; BLINK: \vf{+0.7}).
\uv and the two VisMin baselines move CV-Bench 2D and BLINK the same way (seed 42): by \vf{+1.7} to \vf{+2.2} and \vf{$-2.0$} to \vf{$-5.2$} points (\cref{tab:main}; intervals exclude zero except \vbn's 2D gain).
Over seeds 42--44, \oursmi, \uv, \vbc, and \vbn move them the same way (CV-Bench 2D: \vf{+3.0}, \vf{+2.5}, \vf{+2.3}, and \vf{+2.0}; BLINK: \vf{$-2.2$}, \vf{$-1.9$}, \vf{$-2.0$}, and \vf{$-4.2$}; every interval excludes zero), with no clear change on CV-Bench 3D, MME, or MMVP.
Random pairs and replay alone do too (CV-Bench 2D: \vf{+2.5} and \vf{+1.8}; BLINK: \vf{$-2.6$} and \vf{$-1.5$}; intervals exclude zero except replay alone's BLINK loss, \vf{$-3.0$} to \vf{$+0.2$}).
Split post hoc, CV-Bench 2D's counting questions rise for every arm in \cref{tab:main}, replay alone included (by \vf{2.2}--\vf{3.6} points), and its relation questions by at most \vf{2.1} (\ours), so its gain is not a spatial gain by itself.

\begin{table}[t]
  \caption{\textbf{Repair on unconditioned pairs.}
  Gold pair accuracy by type (blind human labels; 150 pairs per type, 147 for objects).
  \emph{abs.}: negations about absent objects (75 of the negation pairs).
  Seed means as in \cref{tab:main}; replay only: seed 42.}
  \label{tab:gold}
  \centering
  \footnotesize
  \setlength{\tabcolsep}{3.0pt}
  \begin{tabular}{@{}l ccccccc@{}}
    \toprule
    Model & Attr. & Count & Neg. & \emph{abs.} & Obj. & Spat. & All \\
    \midrule
    Text only   & \vf{54.0} & \vf{94.0} & \vf{50.0} & \vf{6.7}  & \vf{70.7} & \vf{68.0} & \vf{67.3} \\
    \base       & \vf{88.0} & \vf{98.0} & \vf{88.7} & \vf{88.0} & \vf{95.9} & \vf{66.7} & \vf{87.4} \\
    Replay only & \vf{88.7} & \vf{98.0} & \vf{83.3} & \vf{78.7} & \vf{95.9} & \vf{60.7} & \vf{85.3} \\
    \vbn        & \vf{88.0} & \vf{98.0} & \vf{85.1} & \vf{83.6} & \vf{95.2} & \vf{68.2} & \vf{86.9} \\
    \vbc        & \vf{85.8} & \vf{97.8} & \vf{79.3} & \vf{82.2} & \vf{91.8} & \vf{67.8} & \vf{84.5} \\
    \uv         & \vf{81.6} & \vf{98.7} & \vf{37.8} & \vf{4.0}  & \vf{90.0} & \vf{96.2} & \vf{80.8} \\
    \midrule
    \ours       & \vf{90.3} & \vf{98.8} & \vf{79.7} & \vf{65.9} & \vf{96.6} & \vf{96.8} & \vf{92.4} \\
    \bottomrule
  \end{tabular}
\end{table}

\subsection{Repair on unconditioned pairs}
\label{sec:gold}
On the gold pairs, sampled regardless of failure and labeled without our committee (\cref{tab:gold}), \ours raises pair accuracy from \vf{87.4\%} to \vf{92.4\%} (five seeds, \vf{91.8}--\vf{92.8\%}; difference \vf{5.0} points, 95\% interval over images \vf{2.5}--\vf{7.5}), fixing \vf{73.8\%} of the base model's errors and breaking \vf{4.9\%} of its correct pairs, whereas replay alone and both VisMin arms do not improve it (\vf{84.5}--\vf{86.9\%}).
Random pairs do about as well (\vf{92.0\%}, seed 42; \vf{$-0.8$} points from \ours's seed 42, interval \vf{$-2.5$} to \vf{$+0.8$}): on gold, selecting failures adds little.
The gain comes from spatial relations (\vf{66.7}$\to$\vf{96.8\%}), which improve this much only in arms trained on our generator's pairs, mined or not (random pairs: \vf{96.0\%}), and it holds where the wording points to the wrong answer (\vf{70.8}$\to$\vf{97.5\%}; interval for the gain \vf{14.6}--\vf{39.6}), so this repair uses the image.
The gain is largest on above/below, the base model's weak axis: its pair accuracy is \vf{91.8\%} on left/right but \vf{42.9\%} on above/below, which \ours raises to \vf{95.6\%} (left/right: \vf{98.1\%}); likewise, two-thirds of the held-out spatial failures are above/below, and \ours repairs \vf{97.6\%} of those against \vf{83.8\%} of the left/right ones.
Gold keeps only pairs that a person answers, with different answers to the two questions, so for spatial relations it keeps \vf{150} of the \vf{372} sampled pairs the author labeled: \vf{153} had a question the author could not answer from the image, \vf{68} got the same answer to both questions, and one exceeded the quota (Supp.~\ref{supp:eval}).
On the other four types together, \ours gains where the wording points to the right answer and loses where it points to the wrong one, ending \vf{1.3} points below the base model.
Negation falls for every trained arm; for \ours it falls by \vf{8.9} points (interval \vf{2.6}--\vf{15.0}), entirely on negations about absent objects (\vf{88.0}$\to$\vf{65.9\%}; replay only: \vf{78.7\%}), which are \vf{6.1\%} of the training negations and half of gold's, while negations about present objects improve (\vf{89.3}$\to$\vf{93.6\%}).
Splitting the labeled answer's log-odds into a blind part and the image's contribution suggests why: on absent-object negations the blind part falls by \vf{10.3} and the image's contribution rises by only \vf{3.9}, so from the wording of negated questions \ours learned the answer that fits present objects (Supp.~\ref{supp:results}).
Random pairs, \vf{23.4\%} of whose training negations ask about the existence of an absent object (\ours: \vf{1.9\%}), keep that kind at the base model's \vf{96.0\%} (\ours, seed 42: \vf{72.0\%}), but fall from \vf{72.0\%} to \vf{16.0\%} on spatial-relation negations about absent objects, which they almost never trained on (\vf{8} of \vf{2{,}400}).
Over all \vf{244} pairs whose wording points to the wrong answer, \ours is therefore no better than the base model (\vf{$-2.5$} points, interval \vf{$-7.6$} to \vf{$+2.9$}; \cref{tab:main}, \emph{Conf.}).
Repair on held-out failures hides this: there, \ours repairs \vf{88.9\%} of negation failures.
These results hold on the \vf{735} gold pairs whose author labels match the stored ones (\ours: \vf{93.7\%}, base model: \vf{88.2\%}; spatial relations: \vf{97.4\%} \vs\ \vf{67.1\%}; Supp.~\ref{supp:results}).

\emph{Beyond our generator.}
The gold test shares our generator's templates and annotation sources, so it cannot show whether the spatial repair carries over to new photos.
Rewriting its spatial questions shows that most of the gain over the base model survives new wording, although \ours's own accuracy falls when its questions are paraphrased: it scores \vf{87.3\%} instead of \vf{96.8\%} (interval for the change \vf{$-14.0$} to \vf{$-5.3$}), still well above the base model's \vf{62.0\%}, losing mostly on above/below, while converse questions, which swap the two objects, cost it \vf{2.3} points, an interval including zero (Supp.~\ref{supp:eval}).
On What'sUp's controlled photos, which the base model nearly solves (\vf{95.6\%} in our yes/no format), \ours falls to \vf{89.4\%} (five seeds; difference \vf{$-6.2$} points, interval \vf{$-8.2$} to \vf{$-4.2$}).
The loss comes from the \vf{103} photos of an object under another, the only relation that declines: \ours answers both questions correctly for \vf{26.4\%} of them (base model: \vf{84.5\%}; change \vf{$-58.1$} points, interval \vf{$-67.3$} to \vf{$-48.9$}), against \vf{91.1\%} of the matching photos with the object on top (base model: \vf{86.4\%}; for this and the other relations, the interval for the change includes zero).
Every arm trained on our pairs loses on the ``under'' photos (\uv: \vf{2.3\%}; random pairs: \vf{45.6\%}), while replay alone and \vbn do not (\vf{88.3\%} and \vf{87.7\%}; Supp.~\ref{supp:results}).
The error is one-sided (post hoc): on these photos \ours affirms the true ``under'' question (\vf{99.8\%}) but also the false ``on'' question (\vf{70.3\%}; base model: \vf{11.7\%}); on ``on'' photos it affirms the false ``under'' question for only \vf{8.7\%}.
It does not need our directional relation pairs: a post hoc seed-42 run trained without them affirms the false ``on'' at least as often (\vf{79.6\%}) and the false ``under'' too (\vf{86.4\%} of ``on'' photos), while the arms trained with them usually reject the false ``under'' but not the false ``on''.
Asked with ``below'' and ``above'', the words of our relation pairs, \ours answers the ``under'' photos correctly (\vf{99.6\%}; base model: \vf{96.1\%}) but affirms ``above'' for only \vf{46.0\%} of the ``on'' photos (base model: \vf{99.0\%}; Supp.~\ref{supp:results}).
With caption choice, What'sUp's own format, this loss does not appear: the base model scores \vf{97.6\%}, and \ours stays at that level (\vf{97.7\%}; interval for the change \vf{$-0.3$} to \vf{$+0.4$}) and picks the right caption for every ``under'' photo, as the base model does; by relation, its only clear changes are a gain on ``behind'' (\vf{91.4\%} to \vf{94.7\%}; interval \vf{$+1.6$} to \vf{$+5.1$}) and a small decline on left/right (\vf{99.0\%} \vs\ \vf{99.5\%}; interval \vf{$-0.9$} to \vf{$-0.2$}).
The other arms evaluated on caption choice also stay at \vf{97.6}--\vf{97.9\%} (three seeds for \uv and the VisMin arms, seed 42 for the others), and each of them that was trained on our pairs still picks the right caption for nearly every ``under'' photo (at least \vf{99.5\%} of option orders) although its yes/no answers about them fail on most (Supp.~\ref{supp:results}).
The model thus still tells ``under'' apart when it chooses among captions; what training breaks is how it answers our yes/no questions about these photos, in ways that depend on the relation word.

\subsection{Why verification matters}
\label{sec:verify-results}
Mined failures carry more label problems than unconditioned pairs: on the sampled gold candidates whose two questions the author answered, the author's answers and the stored labels disagree on \vf{16.6\%} of pairs (\vf{153} of \vf{920}), against \vf{37} wrong labels in the \vf{100} audited mined failures (\cref{sec:verification}).
Removing the committee changes only which mined pairs are trained on.
The unverified arm repairs \vf{76.0\%} (three seeds), \vf{6.8} points below \ours (seed 42 \vs\ seed 42: \vf{6.9} points, 95\% interval over images \vf{4.8}--\vf{8.9}; \ours's seed range: \vf{1.3}).
VAL contains only committee-approved failures, so this comparison favors the distribution \ours was trained on; on the gold test, labeled without the committee, the gap widens: the unverified arm falls below the base model (\vf{80.8\%} \vs\ \vf{87.4\%}; difference \vf{$-6.6$} points, interval \vf{$-10.1$} to \vf{$-3.2$}) and \vf{11.6} points below \ours (\vf{9.3}--\vf{14.0}), breaking \vf{17.6\%} of the pairs the base model gets right.
Most of the loss comes from negation (\vf{68.9\%} \vs\ \vf{88.9\%}), and the blind diagnostic shows why.
On negated questions, the image does not help the unverified model: its sighted--blind gap averages \vf{$-6.0$} points (seeds \vf{$-6.0$}, \vf{$-13.3$}, and \vf{$+1.3$}), whereas \ours's is \vf{$+25.2$} (five seeds).
It has learned to answer negated templates from their text: on gold negations about absent objects it is right \vf{4.0\%} of the time, like the text-only classifiers (\vf{6.7\%}; base model: \vf{88.0\%}; \cref{tab:gold}).
None of the development benchmarks we report sees the shortcut (\uv is within \vf{0.7} points of \ours on each; \cref{tab:main}); among our evaluations only the blind diagnostic and the gold test expose it.

\begin{table}[t]
  \caption{\textbf{Repair accuracy is not a selection criterion.}
  Recipe ablations (seed 42).
  The recipes with the highest repair (terse targets, $w{\ge}6$, little or no replay) have higher blind accuracy or a higher yes-rate and lower precision on RePOPE, but similar gold accuracy.
  \ours uses corrective targets, $w{=}2$, 12k replay rows, and verified failures.}
  \label{tab:ablations}
  \centering
  \footnotesize
  \setlength{\tabcolsep}{3.0pt}
  \begin{tabular}{@{}l ccc ccc c@{}}
    \toprule
    & \multicolumn{3}{c}{Internal (VAL)} & \multicolumn{3}{c}{RePOPE (macro)} & Gold \\
    \cmidrule(lr){2-4}\cmidrule(lr){5-7}\cmidrule(lr){8-8}
    Variant & Repair & Blind & Gap & F1 & Yes & Prec. & Acc. \\
    \midrule
    \base & \vf{10.0} & \vf{1.9} & \vf{8.1} & \vf{93.2} & \vf{44.1} & \vf{92.4} & \vf{87.4} \\
    \ours & \vf{82.8} & \vf{48.3} & \vf{34.5} & \vf{93.0} & \vf{43.6} & \vf{92.7} & \vf{92.8} \\
    \midrule
    \multicolumn{8}{@{}l}{\emph{Target style}} \\
    \quad terse, $w{=}1$       & \vf{84.0} & \vf{50.3} & \vf{33.7} & \vf{88.2} & \vf{51.7} & \vf{81.3} & \vf{93.7} \\
    \quad corrective, $w{=}1$  & \vf{80.5} & \vf{46.1} & \vf{34.5} & \vf{93.2} & \vf{42.0} & \vf{94.6} & \vf{91.8} \\
    \multicolumn{8}{@{}l}{\emph{Decision-token weight}} \\
    \quad $w{=}4$ & \vf{83.7} & \vf{48.1} & \vf{35.6} & \vf{92.5} & \vf{45.1} & \vf{90.7} & \vf{92.8} \\
    \quad $w{=}6$ & \vf{83.9} & \vf{52.5} & \vf{31.3} & \vf{91.6} & \vf{46.1} & \vf{89.0} & \vf{92.6} \\
    \quad $w{=}8$ & \vf{83.9} & \vf{53.4} & \vf{30.5} & \vf{91.2} & \vf{46.7} & \vf{88.0} & \vf{93.3} \\
    \multicolumn{8}{@{}l}{\emph{Replay rows (repair fixed at 12k pairs)}} \\
    \quad 0 (750 updates) & \vf{84.1} & \vf{54.7} & \vf{29.5} & \vf{78.6} & \vf{65.2} & \vf{66.6} & \vf{93.6} \\
    \quad 3k  & \vf{83.5} & \vf{53.9} & \vf{29.6} & \vf{91.2} & \vf{47.1} & \vf{87.7} & \vf{92.2} \\
    \quad 6k  & \vf{82.3} & \vf{50.2} & \vf{32.1} & -- & -- & -- & \vf{91.3} \\
    \quad 24k & \vf{82.7} & \vf{48.2} & \vf{34.5} & \vf{92.7} & \vf{41.6} & \vf{94.6} & \vf{93.3} \\
    \quad 48k & \vf{80.9} & \vf{49.9} & \vf{30.9} & -- & -- & -- & \vf{92.6} \\
    \multicolumn{8}{@{}l}{\emph{Training data}} \\
    \quad unverified failures & \vf{75.9} & \vf{43.7} & \vf{32.2} & \vf{93.2} & \vf{41.6} & \vf{95.1} & \vf{80.7} \\
    \quad random verified pairs & \vf{73.4} & \vf{32.7} & \vf{40.7} & \vf{93.4} & \vf{42.4} & \vf{94.4} & \vf{92.0} \\
    \bottomrule
  \end{tabular}
\end{table}

\subsection{Selecting a recipe}
\label{sec:selection}
\cref{tab:ablations} changes one component at a time (seed 42); Supp.~\cref{fig:selection} plots it with a learning-rate sweep.
At fixed replay, terse \yes/\no targets give the highest repair (\vf{84.0\%}) but make the model answer \yes more often on RePOPE (\vf{44.1}$\to$\vf{51.7\%}), lowering precision from \vf{92.4} to \vf{81.3}; our own registered decision rule, which used repair, chose them.
Corrective targets keep the prior but repair \vf{3.5} points less; at the mean target length the \yes/\no token gets only \vf{9.2\%} of a corrective target's loss weight, and doubling that weight ($w{=}2$) recovers \vf{2.3} points while RePOPE's yes-rate rises by only \vf{1.6}.
Heavier weighting, less replay, and higher learning rates raise the yes-rate or blind accuracy at about the same repair (Supp.~\cref{fig:selection}).
We chose $w{=}2$ and 12k replay rows by judgment (our registered rules preferred $w{=}4$ and 24k rows), so \ours is not the winner of any single registered rule.
Gold accuracy varies little across these recipes (\vf{91.3}--\vf{93.7\%}, terse targets highest; \ours's five seeds: \vf{91.8}--\vf{92.8\%}); they differ more in the answer prior and blind accuracy, which repair accuracy does not reveal (TEST repair: \vf{81.1}--\vf{85.9\%}, in nearly the same order as on VAL).
On the rephrased VAL, terse targets lose only \vf{1.8} points and $w{=}8$ \vf{3.5}, although both also lower RePOPE precision; every other variant or arm we tested on it that trains on our pairs (seed 42) loses \vf{5.3}--\vf{9.1}, within the range of \ours's three seeds (\vf{5.1}--\vf{9.7}; Supp.~\ref{supp:training}).

\begin{table}[t]
  \caption{\textbf{A side effect of training and its partial fix.}
  VisMin relation, two-image selection (622 groups).
  \emph{Paired}: both opposite captions right; \emph{Sens.}: different images chosen for the two captions; \emph{Cond.}: paired accuracy among such groups; \emph{Tok.}: median output tokens; MI: multi-image replay (seeds 43--46: range).
  NegVQA: negated questions (seed 42).}
  \label{tab:sideeffect}
  \centering
  \footnotesize
  \setlength{\tabcolsep}{3.2pt}
  \begin{tabular}{@{}l cccc c@{}}
    \toprule
    & \multicolumn{4}{c}{VisMin relation (IMAGE)} & NegVQA \\
    \cmidrule(lr){2-5}\cmidrule(lr){6-6}
    Model & Paired & Sens. & Cond. & Tok. & negated \\
    \midrule
    \base                          & \vf{74.3} & \vf{77.2} & \vf{96.3} & \vf{131} & \vf{62.3} \\
    \quad + describe first         & \vf{72.5} & \vf{74.4} & \vf{97.4} & \vf{189} & -- \\
    \ours, seed 42                 & \vf{47.1} & \vf{48.6} & \vf{97.0} & \vf{2}   & \vf{69.6} \\
    \quad + describe first         & \vf{62.5} & \vf{64.6} & \vf{96.8} & \vf{165} & -- \\
    \ours, 5-seed mean             & \vf{49.6} & \vf{51.4} & \vf{96.5} & \vf{2}   & -- \\
    \uv, seed 42                   & \vf{46.8} & \vf{48.4} & \vf{96.7} & \vf{2}   & \vf{69.3} \\
    Random pairs                   & \vf{46.5} & \vf{48.1} & \vf{96.7} & \vf{2}   & -- \\
    Replay only                    & \vf{48.4} & \vf{50.0} & \vf{96.8} & \vf{2}   & -- \\
    \midrule
    \oursmi, seed 42               & \vf{76.0} & \vf{78.1} & \vf{97.3} & \vf{126} & \vf{68.3} \\
    \quad seeds 43--46             & \vf{55.9}--\vf{60.8} & \vf{58.2}--\vf{63.2} & \vf{95.9}--\vf{96.2} & \vf{2}--\vf{3} & -- \\
    \quad 5-seed mean              & \vf{62.0} & \vf{64.3} & \vf{96.3} & --       & -- \\
    \bottomrule
  \end{tabular}
\end{table}

\subsection{A side effect on two-image questions}
\label{sec:sideeffect}
Training also changes how the model handles a task it was not trained on.
VisMin's relation subset~\cite{vismin} shows two images that differ in one spatial relation and asks which image matches a caption, once for each of two opposite captions (622 groups).
The base model describes both images before answering (median \vf{131} tokens) and gets both captions right in \vf{74.3\%} of groups.
After training, \ours answers with a bare ``First'' or ``Second'' (median \vf{2} tokens), and paired accuracy falls to \vf{49.6\%} (five-seed mean; every seed \vf{20.9}--\vf{27.2} points below the base model).

Among groups in which the model picks different images for the two captions, it is right \vf{96}--\vf{97\%} of the time before and after training (\cref{tab:sideeffect}, \emph{Cond.}); what falls is caption sensitivity: the trained model picks the same image for both captions in about half of the groups, including \vf{206} of the \vf{209} groups that the base model gets right and \ours (seed 42) gets wrong (by relation axis: Supp.~\ref{supp:sideeffect}).
Asking the trained model to describe each image first recovers \vf{15.4} points net of new errors for seed 42 (five seeds: \vf{6.1}--\vf{15.4}, mean \vf{10.7}), regaining \vf{123} of those \vf{209} groups, whereas the same prompt slightly lowers the base model's paired accuracy (\vf{74.3\%} to \vf{72.5\%}): describing first is not simply a better prompt; it undoes part of what training broke.
Held-out BLINK points the same way: on its four two-image tasks \ours is \vf{10.8} points below the base model on average (replay alone: \vf{7.3}), against \vf{0.4} above on single-image tasks and \vf{1.4} below on tasks with three or four images (no registered intervals at this level).
The unverified arm drops in the same way, as do random pairs (\vf{46.5\%}) and replay alone (\vf{48.4\%}, median \vf{2} tokens), so the repair data are not needed for the drop.

\emph{Multi-image replay.}
We replaced \vf{1{,}500} single-image description rows of the replay data with \vf{750} two-image caption-selection groups built from unedited COCO training photos, with the base model's own responses as targets, kept only when it chose correctly for both captions.
In the pre-registered seed-42 run, this recovered paired accuracy fully (\vf{76.0\%}, \vs\ \vf{74.3\%} for the base model), at a repair cost of \vf{0.5} points.
The replication was weaker: over five matched seeds, multi-image replay gains \vf{28.9}, \vf{9.8}, \vf{8.5}, \vf{12.1}, and \vf{2.6} points over \ours, a mean of \vf{12.4} (95\% $t$-interval over seeds: \vf{0.1} to \vf{24.7}), all positive but only seeds 42 and 45 above the registered 10-point criterion.
This recovers about half of \ours's average loss at essentially no repair cost (\vf{82.8\%} on VAL with and without it, five-seed means).
Recovery tracks whether the model resumes describing before choosing: seed 42 answers with a median of \vf{126} tokens and recovers fully, while seeds 43--46 stay terse (\vf{2}--\vf{3} tokens) and recover partly, mainly through higher caption sensitivity.
The other costs are small: \vf{0.4} points on gold on average (seed 42: \vf{2.8}, concentrated on negation).

\section{Limitations}
\label{sec:limitations}
We repair one model; mining is model-specific, and a second backbone is future work.
Questions come from templates: much of VAL and two-thirds of gold are predictable from the question text alone (\cref{sec:main}), failures on free-form questions are not tested, and the gold test shares the generator; its rewritten spatial questions change only the wording and the order of the objects.
The committee is a filter, not ground truth: on the audited failures it discards a third of the model errors (\cref{sec:verification}), and \vf{5.0\%} of verified target cells were factually wrong, including \vf{5} of \vf{40} counting cells, whose labels come from captions.
All human labels, including both audits and the gold test, come from one of the authors, without a second annotator or adjudication; on the \vf{200} questions shown twice during gold labeling, the two answers agreed \vf{194} times.
VAL and TEST contain only committee-approved failures, VAL and the development benchmarks informed our decisions, the What'sUp and rewritten-gold tests were specified after the main results, and most ablations, including the random-pairs control, have one seed.
The VisMin baselines are our reconstruction on our model and recipe.
The multi-image fix is partial and varies across seeds, and outside VisMin the side effect shows only in BLINK's subtasks.
Finally, the blind diagnostic uses an unusual input, a black image, and the LLM extractor was validated on samples; unresolved answers count as wrong.

\section{Conclusion}
\label{sec:conclusion}
Training a VLM on its own mined failures can repair real errors, here spatial relations (\vf{66.7\%} to \vf{96.8\%} on independently labeled pairs), even where the question's wording points to the wrong answer, but repair accuracy on held-out failures overstates what it repairs.
Four lessons apply beyond our setting.
First, verify mined failures: they carry more label problems than unconditioned pairs (\cref{sec:verify-results}), and training on them taught a negation shortcut that pushed accuracy on unconditioned pairs below the base model, unseen by our development benchmarks.
Second, measure repair on independently labeled, unconditioned pairs, separated by whether the question's wording alone gives the answer, because held-out failures share the wording cues of the training failures and rarely contain what training breaks; selecting failures added \vf{8.5} points of TEST repair but \vf{0.8} on gold.
Third, test tasks the training did not target: two-image relation accuracy fell by \vf{24.6} points (five-seed mean), and BLINK's two-image tasks by \vf{10.8} (seed 42), regressions that no single-image benchmark can show.
Fourth, test a repair beyond the generator that produced it: rewording the questions cost \vf{16.1} points of spatial repair on held-out failures and \vf{9.5} points of accuracy on gold spatial pairs, where \ours stayed well above the base model (\vf{87.3\%} \vs\ \vf{62.0\%}), and on new photos the model's answers to our yes/no questions about objects under others collapsed while its caption choices for the same photos stayed perfect.

{
    \small
    \bibliographystyle{ieeenat_fullname}
    \bibliography{refs}
}

\ifsupp
  \clearpage
\ifdefined\arxivversion\else\setcounter{page}{1}\fi %
\maketitlesupplementary
\setcounter{section}{0}
\renewcommand{\thesection}{\Alph{section}}
\renewcommand{\theHsection}{supp.\Alph{section}} %

\section{Typed failure mining}
\label{supp:mining}

\emph{Pairs.}
\cref{tab:pairexamples} shows the first TRAIN pair of each type.
Spatial pairs use four relations in the image frame: to the left of, to the right of, above, and below.
Counting pairs take the true number from a COCO caption, matched by the instance annotations (\vf{1}--\vf{9} in TRAIN), and ask about a different number (\vf{0}--\vf{13} in TRAIN).
Negation pairs negate an existence, attribute, or relation fact about a present or an absent object (six operators).
TRAIN has \vf{2{,}254} negation pairs about present objects (existence \vf{1{,}268}, relation \vf{956}, attribute \vf{30}) and only \vf{146} about absent ones (\vf{46}, \vf{85}, \vf{15}); VAL has \vf{281} and \vf{19}, and gold \vf{25} for each operator.
\begin{table*}[t]
  \caption{\textbf{Typed minimal pairs.} The first TRAIN pair of each type, in file order.}
  \label{tab:pairexamples}
  \centering
  \footnotesize
  \setlength{\tabcolsep}{4pt}
  \begin{tabular}{@{}l l p{0.34\textwidth} p{0.34\textwidth}@{}}
    \toprule
    Type & Source & Question 1 (stored answer) & Question 2 (stored answer) \\
    \midrule
    Object existence & GQA & Is there a lamp in the image? (\yes) & Is there a picture in the image? (\no) \\
    Attribute & GQA & Is there a white drape in the image? (\yes) & Is there a purple drape in the image? (\no) \\
    Spatial relation & COCO & In the image, is there a car above a skateboard? (\yes) & In the image, is there a car below a skateboard? (\no) \\
    Counting & COCO captions & Is there exactly one toilet in the image? (\yes) & Are there exactly six toilets in the image? (\no) \\
    Negation & COCO & In the image, is there a toaster to the left of a knife? (\yes) & In the image, is there no toaster to the left of a knife? (\no) \\
    \bottomrule
  \end{tabular}
\end{table*}

\emph{Wrappers and scoring.}
Each prompt shows the image first and then one of three instructions, followed by a new line and ``Question: \{question\}'', inside the model's chat template.
Mining uses ``Answer the question using only one word: Yes or No.''; evaluation (VAL, TEST, gold, and their rewritten versions) uses ``Answer the question based only on what is visible in the image. Answer Yes or No.''; and training uses ``Answer the question based only on what is visible in the image. Answer Yes or No, then briefly explain in one sentence.''
A cell is scored by one forward pass over the prompt, comparing the log probabilities of the single tokens \yes and \no at the first answer position; an absolute difference of at most $10^{-8}$ is a tie.
Cells are scored one at a time; Supp.~\ref{supp:eval} lists the pinned software.

\emph{Mining.}
A pair is a mined failure if both of its cells are resolved and at least one disagrees with its stored label; there is no margin threshold, and pairs with a tie are excluded.
\cref{tab:mining} gives the failure rates by type.
Round two generated \vf{500{,}000} new pairs per type, excluding round one's; a seeded quota drew \vf{20{,}000} of its failures per type, from which batches went to the committee until each type had at least \vf{3{,}000} kept pairs (Supp.~\ref{supp:committee}).
\begin{table}[t]
  \caption{\textbf{Mined failures of the base model.} Share of the scored pairs that failed (\%); pairs with a tie (round one: \vf{1.6\%}; round two: \vf{1.2\%}) count in the denominator but not as failures.}
  \label{tab:mining}
  \centering
  \footnotesize
  \setlength{\tabcolsep}{4pt}
  \begin{tabular}{@{}l rr rr@{}}
    \toprule
    & \multicolumn{2}{c}{Round one} & \multicolumn{2}{c}{Round two} \\
    \cmidrule(lr){2-3}\cmidrule(lr){4-5}
    Type & Pairs & Failed & Pairs & Failed \\
    \midrule
    Object existence & \vf{20{,}000} & \vf{13.2} & \vf{500{,}000} & \vf{14.2} \\
    Attribute        & \vf{15{,}000} & \vf{25.7} & \vf{500{,}000} & \vf{25.9} \\
    Spatial relation & \vf{15{,}000} & \vf{43.9} & \vf{500{,}000} & \vf{53.7} \\
    Counting         & \vf{4{,}000}  & \vf{6.3}  & \vf{500{,}000} & \vf{6.7} \\
    Negation         & \vf{15{,}000} & \vf{27.3} & \vf{500{,}000} & \vf{10.7} \\
    \midrule
    All & \vf{69{,}000} & \vf{25.3} & \vf{2{,}500{,}000} & \vf{22.2} \\
    \bottomrule
  \end{tabular}
\end{table}

\emph{Why the base model passes \vf{10\%} of VAL.}
Every VAL pair failed at mining, yet under the evaluation wrapper the base model passes \vf{150} of the \vf{1{,}500} (spatial relations: \vf{23} of \vf{300}).
At mining, the \vf{153} cells that later flipped had favored the wrong answer by a median log-probability difference of \vf{0.5}, and a tenth of them by \vf{1.75} or more, so they were not all near-ties; the change of wrapper is the likeliest cause, but no controlled comparison isolates it.
Rescored with the mining wrapper, under which the base model fails every VAL pair, \ours (seed 42) repairs \vf{81.2\%} (evaluation wrapper: \vf{82.8\%}).

\emph{Splits.}
The \vf{15{,}000} verified pairs (\vf{5{,}974} from round one, \vf{9{,}026} from round two) were assigned to TRAIN, VAL, and TEST (\vf{2{,}400}, \vf{300}, and \vf{300} per type) by whole image-identity groups, which join images transitively by source ID, path, SHA-256, and perceptual hash (seed 20260812): \vf{9{,}528}, \vf{1{,}336}, and \vf{1{,}330} groups, with no collision across splits.

\section{Committee verification}
\label{supp:committee}

\emph{Judges and prompt.}
The judges are called as \texttt{claude-sonnet-5}, a Seed2.1~Pro deployment endpoint, and \texttt{gemini-3.6-flash}; the providers' fixed model revisions are not recorded.
Each judge receives the image and one question at a time with the prompt below, and no stored label, model answer, score, or other judge's answer:
\begin{quote}\small\ttfamily\raggedright
Inspect the image and answer the question using only visible image evidence.\\[2pt]
Return exactly one of:\\ yes\\ no\\ uncertain\\[2pt]
Use "uncertain" when the image does not provide enough visible evidence, the question is ambiguous, or neither yes nor no can be determined reliably.\\[2pt]
Return no explanation and no additional text.\\[2pt]
Question:\\ \{question\}
\end{quote}

\emph{Decisions.}
A two-of-three majority answers each question.
With three valid answers, a pair is kept if both majorities agree with the stored labels, a label disagreement if both questions have a \emph{yes}/\emph{no} majority and at least one contradicts its label, and ambiguous if a question has an \emph{uncertain} majority or none; pairs with a missing or malformed answer go to manual review.
\cref{tab:committeetypes} gives the outcomes by type, and \cref{fig:pipeline} shows a rejected pair.
The committee judged all \vf{17{,}425} round-one failures (keeping \vf{34.3\%}) and \vf{28{,}005} round-two failures (\vf{32.4\%}); the verified pool takes \vf{15{,}000} of the \vf{15{,}059} kept pairs to fill its per-type quotas.
Over the \vf{90{,}833} questions with three valid answers, all three judges agree on \vf{61.7\%}, least on spatial relations (\vf{52.8\%}) and most on object existence (\vf{68.6\%}); pairwise agreement is \vf{69.3}--\vf{76.6\%} (raw rates, \emph{uncertain} counted as an answer).
\begin{table}[t]
  \caption{\textbf{Committee outcomes by type.} Mined failures judged in both rounds (\% of judged pairs); \vf{15} pairs with a missing or malformed answer went to manual review.}
  \label{tab:committeetypes}
  \centering
  \footnotesize
  \setlength{\tabcolsep}{4pt}
  \begin{tabular}{@{}l r rrr@{}}
    \toprule
    Type & Judged & Kept & Label disagr. & Ambiguous \\
    \midrule
    Object existence & \vf{11{,}890} & \vf{25.2} & \vf{66.9} & \vf{7.8} \\
    Attribute        & \vf{8{,}604}  & \vf{35.0} & \vf{56.4} & \vf{8.6} \\
    Spatial relation & \vf{8{,}837}  & \vf{34.0} & \vf{55.5} & \vf{10.5} \\
    Counting         & \vf{8{,}454}  & \vf{35.7} & \vf{53.5} & \vf{10.7} \\
    Negation         & \vf{7{,}645}  & \vf{39.5} & \vf{52.7} & \vf{7.8} \\
    \midrule
    All & \vf{45{,}430} & \vf{33.1} & \vf{57.8} & \vf{9.0} \\
    \bottomrule
  \end{tabular}
\end{table}

\emph{Human audits.}
In the audit of \vf{100} mined failures before verification, wrong labels were most frequent for attributes (\vf{10} of \vf{20}) and object existence (\vf{9}), then spatial relations (\vf{8}), counting, and negation (\vf{5} each).
The committee of that time, with Gemini~3.5~Flash as its third judge, kept the model errors with \vf{80\%} precision (\vf{32} of \vf{40}) and \vf{71\%} recall (\vf{32} of \vf{45}).
\cref{tab:committee100} compares the final committee with the author's classes on the \vf{100} audited failures, which were its calibration set (\cref{sec:verification}).
It keeps \vf{35} of them, \vf{30} of which are model errors, and discards \vf{15} of the \vf{45} model errors, \vf{12} of them as label disagreements; of the \vf{37} wrong labels it keeps \vf{1}.
\begin{table}[t]
  \caption{\textbf{The final committee on its calibration set.}
  The \vf{100} audited mined failures (\vf{20} per type): it keeps model errors with \vf{85.7\%} precision (\vf{30} of \vf{35}) and \vf{66.7\%} recall (\vf{30} of \vf{45}).}
  \label{tab:committee100}
  \centering
  \footnotesize
  \setlength{\tabcolsep}{4.0pt}
  \begin{tabular}{@{}l c ccc@{}}
    \toprule
    & & \multicolumn{3}{c}{Committee} \\
    \cmidrule(lr){3-5}
    Author's class & $n$ & Kept & Label disagr. & Ambiguous \\
    \midrule
    Model error & \vf{45}  & \vf{30} & \vf{12} & \vf{3} \\
    Wrong label & \vf{37}  & \vf{1}  & \vf{35} & \vf{1} \\
    Ambiguous   & \vf{18}  & \vf{4}  & \vf{10} & \vf{4} \\
    \midrule
    All         & \vf{100} & \vf{35} & \vf{57} & \vf{8} \\
    \bottomrule
  \end{tabular}
\end{table}

\emph{The controls' training pairs.}
Of the \vf{12{,}000} pairs \uv trains on, \vf{3{,}202} had also gone to the committee in round two: \vf{1{,}996} of these (\vf{62.3\%}) were label disagreements, \vf{900} were kept, \vf{304} were ambiguous, and \vf{2} went to review.
The other \vf{8{,}798} were never judged, so we do not extrapolate a rate.
For random pairs, the committee judged \vf{20{,}695} pairs drawn regardless of failure and kept \vf{72.5\%}, from \vf{49.7\%} of the spatial pairs to \vf{87.0\%} of the counting pairs.

\section{Corrective targets and human audit}
\label{supp:targets}

\emph{Templates.}
Each type has two template families with four surface variants each (negation: four families of four variants plus three variants for overlap, \vf{19} in all), filled from the pair's stored slots, such as the object, the true and the asked color or number, or the relation (\cref{tab:targets}).
Every target begins with \yes or \no as a single token, the token that first-token scoring compares, so a corrective target trains the scored decision directly and states the fact after it.
\begin{table*}[t]
  \caption{\textbf{Corrective targets.} Targets of the pairs in \cref{tab:pairexamples}, with the number of templates per type and the mean target length in TRAIN (tokens, without the end-of-turn token).}
  \label{tab:targets}
  \centering
  \footnotesize
  \setlength{\tabcolsep}{4pt}
  \begin{tabular}{@{}l c c p{0.28\textwidth} p{0.28\textwidth}@{}}
    \toprule
    Type & Templates & Tokens & Target for question 1 & Target for question 2 \\
    \midrule
    Object existence & \vf{8}  & \vf{8.7}  & Yes, there is a lamp in the image. & No, there is no picture in the image. \\
    Attribute        & \vf{8}  & \vf{10.0} & Yes, a drape appears white. & No, a drape is white rather than purple. \\
    Spatial relation & \vf{8}  & \vf{14.7} & Yes, a car is above a skateboard. & No, in the image, a car is above a skateboard, not below it. \\
    Counting         & \vf{8}  & \vf{10.1} & Yes, the image shows one toilet. & No, there is one toilet, not six. \\
    Negation         & \vf{19} & \vf{10.8} & Yes, in the image, a toaster is to the left of a knife. & No, the position of a toaster is to the left of a knife. \\
    \midrule
    All & & \vf{10.87} & & \\
    \bottomrule
  \end{tabular}
\end{table*}

\emph{Decision weight.}
In a target of length $L$, the decision token's share of the cell's loss weights is $w/(w+L-1)$ (\cref{eq:cellloss}); at the mean length of \vf{10.87} tokens it is \vf{9.2\%}, \vf{16.9\%}, \vf{28.8\%}, \vf{37.8\%}, and \vf{44.8\%} for $w = 1$, 2, 4, 6, and 8 (\cref{tab:ablations}).
Replay rows are never weighted this way.

\emph{Audit of the rendered targets.}
The author checked \vf{100} TRAIN pairs (\vf{20} per type, \vf{200} target cells; seed 20261020) against their images.
A cell's target counts as correct if its decision and fact are right and its wording is acceptable: \vf{176} of the \vf{200} were (\vf{81} of \vf{100} pairs).
\vf{10} stated a wrong fact or decision, \vf{5} of them in counting, whose true numbers come from captions; spatial relations had \vf{3}, object existence \vf{2}, and attributes and negation none.
The remaining verdicts were \vf{12} uncertain and \vf{2} ambiguous questions.

\section{Replay data}
\label{supp:replay}
\cref{tab:replay} lists the replay rows by category and target source.
They are the first \vf{12{,}000} of a seeded ranking of a \vf{60{,}000}-row replay release, so the larger and smaller replay sets of \cref{tab:ablations} are nested.
Descriptions, open-ended VQA, and text-only instructions take the frozen base model's own greedy response as target (bf16, seed 42, batches of 4; at most \vf{128} new tokens for open-ended VQA and \vf{256} for the others); replay keeps each source's own instruction, without a repair wrapper.
Replay images exclude every image of the frozen splits, and held-out benchmark items whose images occur in the replay data are left out of the clean populations (Supp.~\ref{supp:eval}).
\begin{table}[t]
  \caption{\textbf{Replay data.} The \vf{12{,}000} replay rows by category, with the source of their targets; \vf{8{,}020} targets come from the base model.}
  \label{tab:replay}
  \centering
  \footnotesize
  \setlength{\tabcolsep}{4pt}
  \begin{tabular}{@{}l l r l@{}}
    \toprule
    Category & Source & Rows & Target \\
    \midrule
    Detailed descriptions & LLaVA, COCO & \vf{3{,}961} & base model \\
    Open-ended VQA & LLaVA, VQAv2 & \vf{3{,}143} & base model \\
    Yes/no VQA & VQAv2 (balanced) & \vf{2{,}884} & dataset \\
    Text-only instructions & Dolly & \vf{916} & base model \\
    Chart QA & ChartQA & \vf{642} & dataset \\
    Science QA & ScienceQA & \vf{379} & dataset \\
    Counting yes/no & -- & \vf{41} & dataset \\
    Negation yes/no & -- & \vf{34} & dataset \\
    \midrule
    All & & \vf{12{,}000} & \\
    \bottomrule
  \end{tabular}
\end{table}

\section{Training details and recipe sweeps}
\label{supp:training}

\emph{Training settings.}
\cref{tab:hparams} lists the training settings of \ours.
In an earlier registered experiment we also supervised the end-of-turn token, and we kept it unsupervised; over its two seeds, VAL pair accuracy without that supervision had a sample standard deviation of \vf{1.04} points.
A later learning-rate and batch-size sweep (seed 42) found nothing to adopt: the best setting within our tolerances, learning rate $10^{-4}$ at the same batch size, gained \vf{0.39} points on the registered development objective, below the registered threshold of \vf{0.82} points (the mean range of \ours's three seeds on the four development metrics), so the reference recipe stayed.
\begin{table}[t]
  \caption{\textbf{Training settings.} The settings of \ours; the baselines change only their data.}
  \label{tab:hparams}
  \centering
  \footnotesize
  \setlength{\tabcolsep}{4pt}
  \begin{tabular}{@{}l p{0.66\columnwidth}@{}}
    \toprule
    Setting & Value \\
    \midrule
    Model & \qwen, pinned revision, bf16 \\
    LoRA & rank 32, $\alpha{=}64$, dropout 0.05 \\
    LoRA modules & q, k, v, o, gate, up, and down projections of all 36 language-model layers \\
    Trainable & \vf{87.3}M of \vf{8.85}B parameters (\vf{1.0\%}) \\
    Frozen & vision encoder, projector, base weights, LM head \\
    Optimizer & AdamW, $\beta{=}(0.9, 0.999)$, $\epsilon{=}10^{-8}$, no weight decay \\
    Learning rate & $5{\times}10^{-5}$; 45 warm-up updates, then cosine decay to 0 \\
    Gradient clipping & global norm 1.0 \\
    Batch & 8 repair pairs + 8 replay rows (micro-batches of 2 + 2) \\
    Updates & 1,500 (one epoch) \\
    Loss & $\frac12$ repair + $\frac12$ replay; $w{=}2$; end of turn unsupervised \\
    Sequence limit & 32,768 tokens with image tokens; longer inputs fail, none are cut \\
    Hardware & one GPU, listed as an RTX 6000 Ada (48\,GB) in a later inventory; \vf{4.1} h for the training loop \\
    \bottomrule
  \end{tabular}
\end{table}

\begin{figure}[t]
  \centering
  \includegraphics[width=\linewidth]{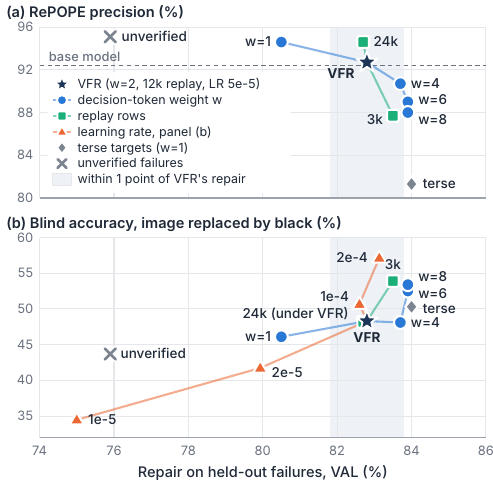}
  \caption{\textbf{Recipes with nearly equal repair differ in what they learned.}
  Seed-42 checkpoints of \cref{tab:ablations} with RePOPE results and, in (b), a learning-rate sweep; lines join variants that change one factor of \ours.
  (a)~The recipes with the highest repair are the least precise on RePOPE.
  (b)~Within one point of \ours's repair (shaded), blind (image-removed) accuracy ranges from \vf{48.1} to \vf{57.1\%}.}
  \label{fig:selection}
\end{figure}
\emph{Rephrased VAL by recipe.}
The rephrased VAL (Supp.~\ref{supp:eval}), evaluated after the recipe was frozen, shows how much each recipe's repair depends on the template wording (\cref{tab:rephrased}).
Terse targets, which supervise only the answer token of each repair target, lose only \vf{1.8} points (spatial relations: \vf{2.3}) and $w{=}8$ \vf{3.5} (\vf{6.3}), although both also lower RePOPE precision (to \vf{81.3} and \vf{88.0}; base model: \vf{92.4}); every other variant or arm in \cref{tab:rephrased} trained on our pairs (seed 42) loses \vf{5.3}--\vf{9.1} points (spatial relations: \vf{12.0}--\vf{21.7}), about as much as \ours's three seeds (\vf{5.1}--\vf{9.7}; spatial relations: \vf{11.0}--\vf{20.0}), while the arms that do not learn our pairs gain overall, like the base model.
\begin{table*}[t]
  \caption{\textbf{Repair on the original and the rephrased VAL} (sighted pair accuracy, \%; changes in points).
  The rephrased VAL keeps each pair's images and labels and rewords its questions (Supp.~\ref{supp:eval}).
  Paired intervals over images exclude zero for every change except \vbn's on spatial relations.}
  \label{tab:rephrased}
  \centering
  \footnotesize
  \setlength{\tabcolsep}{5pt}
  \begin{tabular}{@{}l ccc ccc@{}}
    \toprule
    & \multicolumn{3}{c}{All types (\vf{1{,}500} pairs)} & \multicolumn{3}{c}{Spatial relations (\vf{300} pairs)} \\
    \cmidrule(lr){2-4}\cmidrule(lr){5-7}
    Variant & Original & Rephrased & Change & Original & Rephrased & Change \\
    \midrule
    \base                      & \vf{10.0} & \vf{25.1} & \vf{$+15.1$} & \vf{7.7}  & \vf{14.3} & \vf{$+6.7$} \\
    \ours, seed 42             & \vf{82.8} & \vf{75.3} & \vf{$-7.5$}  & \vf{93.3} & \vf{76.0} & \vf{$-17.3$} \\
    \quad seed 43              & \vf{82.1} & \vf{77.1} & \vf{$-5.1$}  & \vf{92.0} & \vf{81.0} & \vf{$-11.0$} \\
    \quad seed 44              & \vf{83.5} & \vf{73.8} & \vf{$-9.7$}  & \vf{93.3} & \vf{73.3} & \vf{$-20.0$} \\
    \quad mean                 & \vf{82.8} & \vf{75.4} & \vf{$-7.4$}  & \vf{92.9} & \vf{76.8} & \vf{$-16.1$} \\
    \midrule
    \multicolumn{7}{@{}l}{\emph{Recipe variants of \cref{tab:ablations} (seed 42)}} \\
    \quad terse, $w{=}1$       & \vf{84.0} & \vf{82.2} & \vf{$-1.8$}  & \vf{95.0} & \vf{92.7} & \vf{$-2.3$} \\
    \quad corrective, $w{=}1$  & \vf{80.5} & \vf{72.1} & \vf{$-8.4$}  & \vf{92.3} & \vf{71.3} & \vf{$-21.0$} \\
    \quad $w{=}4$              & \vf{83.7} & \vf{78.2} & \vf{$-5.5$}  & \vf{94.3} & \vf{82.0} & \vf{$-12.3$} \\
    \quad $w{=}6$              & \vf{83.9} & \vf{78.6} & \vf{$-5.3$}  & \vf{94.3} & \vf{82.3} & \vf{$-12.0$} \\
    \quad $w{=}8$              & \vf{83.9} & \vf{80.3} & \vf{$-3.5$}  & \vf{94.3} & \vf{88.0} & \vf{$-6.3$} \\
    \quad 3k replay rows       & \vf{83.5} & \vf{76.7} & \vf{$-6.8$}  & \vf{93.3} & \vf{79.0} & \vf{$-14.3$} \\
    \quad 24k replay rows      & \vf{82.7} & \vf{74.5} & \vf{$-8.1$}  & \vf{93.3} & \vf{73.3} & \vf{$-20.0$} \\
    \multicolumn{7}{@{}l}{\emph{Other arms (seed 42)}} \\
    \quad \uv                  & \vf{75.9} & \vf{66.9} & \vf{$-9.1$}  & \vf{92.3} & \vf{70.7} & \vf{$-21.7$} \\
    \quad Random pairs         & \vf{73.4} & \vf{66.3} & \vf{$-7.1$}  & \vf{92.0} & \vf{72.3} & \vf{$-19.7$} \\
    \quad \oursmi              & \vf{82.3} & \vf{74.6} & \vf{$-7.7$}  & \vf{93.3} & \vf{74.0} & \vf{$-19.3$} \\
    \quad \vbc                 & \vf{26.7} & \vf{30.9} & \vf{$+4.2$}  & \vf{31.0} & \vf{23.7} & \vf{$-7.3$} \\
    \quad \vbn                 & \vf{14.9} & \vf{27.2} & \vf{$+12.3$} & \vf{23.0} & \vf{21.7} & \vf{$-1.3$} \\
    \quad Replay only          & \vf{9.9}  & \vf{25.2} & \vf{$+15.3$} & \vf{7.7}  & \vf{13.7} & \vf{$+6.0$} \\
    \bottomrule
  \end{tabular}
\end{table*}

\section{Evaluation protocols}
\label{supp:eval}

\emph{Pinned software.}
All scoring uses \qwen at revision \texttt{0c351dd0} with Transformers 4.57.6 in bf16.
The benchmarks run in VLMEvalKit at commit \texttt{aec2f4e9}; HallusionBench, AMBER, RePOPE, VisMin, and NegVQA use their authors' scorers at pinned commits, and the definitions of MME, MMVP, BLINK, and CV-Bench come from pinned snapshots of their authors' repositories.

\emph{Frozen TEST.}
Before any training, TEST was read for the release checks of the frozen splits, an audit of their wording cues, and a registered diagnostic of the unadapted base model with the real image, a black image, and noise; later work used only its image identities, to keep them out of other data.
No recorded selection of an adapter or a recipe used TEST.

\emph{Gold construction.}
Gold pairs are sampled from all \vf{2{,}569{,}000} generated pairs of both rounds, regardless of failure or committee outcome, after removing every pair whose image group touches our splits, replay data, other training data, the VisMin training release, or the benchmarks: \vf{904{,}542} pairs in \vf{66{,}575} image groups remain.
The author labeled a first sample of \vf{200} pairs per type (seed 20261021; negation balanced between the two answer orders and over its six operators) and then reserves as needed, in an interface that hid the stored labels and model answers, kept the two questions of a pair apart, and repeated \vf{200} questions.
Gold labels are the author's answers; no adjudication or stored label fills them in.
The frozen set has \vf{747} pairs (\vf{150} per type, \vf{147} for object existence); negation has \vf{25} pairs per operator, and half of all questions have the answer \yes.

\emph{Gold labeling outcomes.}
Of the \vf{1{,}276} pairs the author labeled, we dropped, by the first matching rule, \vf{359} with a non-binary (``can't tell'') answer, \vf{1} with an inconsistent answer to a repeated item, and \vf{126} with the same answer to both questions; \vf{43} more fell to other exclusions and the per-type quotas, which leaves \vf{747} (\cref{tab:golddrops}; \vf{64} reserve pairs were never shown).
Spatial pairs are most often unanswerable: the author could not answer \vf{153} of \vf{372} (\vf{41.1\%}), against \vf{15.0}--\vf{29.6\%} of the other types' pairs.
On the \vf{920} pairs with two binary answers, the author's answers and the stored labels disagree on \vf{153} (\vf{16.6\%}; Wilson 95\% interval \vf{14.4}--\vf{19.2}) and on \vf{193} of \vf{2{,}083} binary answers (\vf{9.3\%}); in the frozen set, \vf{12} of the \vf{747} pairs (\vf{1.6\%}) have both stored labels reversed, a rate for the selected pairs, not for the generator.
These are disagreements, not label errors, since either side may be wrong, and a ``can't tell'' answer is not counted as one.
Of the \vf{200} questions shown twice, the two answers agree on \vf{194} (\vf{97.0\%}).
\begin{table}[t]
  \caption{\textbf{Gold pairs labeled and dropped}, by the first matching rule.
  \emph{Other}: one inconsistent repeat (object existence), two identity or duplicate exclusions, and \vf{41} pairs beyond the per-type quotas.}
  \label{tab:golddrops}
  \centering
  \footnotesize
  \setlength{\tabcolsep}{4.0pt}
  \begin{tabular}{@{}l ccccc@{}}
    \toprule
    Type & Labeled & Can't tell & Same answer & Other & Kept \\
    \midrule
    Attribute & \vf{250}       & \vf{74}  & \vf{26}  & \vf{0}  & \vf{150} \\
    Counting  & \vf{200}       & \vf{30}  & \vf{5}   & \vf{15} & \vf{150} \\
    Negation  & \vf{234}       & \vf{50}  & \vf{8}   & \vf{26} & \vf{150} \\
    Object    & \vf{220}       & \vf{52}  & \vf{19}  & \vf{2}  & \vf{147} \\
    Spatial   & \vf{372}       & \vf{153} & \vf{68}  & \vf{1}  & \vf{150} \\
    \midrule
    All       & \vf{1{,}276}   & \vf{359} & \vf{126} & \vf{44} & \vf{747} \\
    \bottomrule
  \end{tabular}
\end{table}

\emph{Rephrased VAL.}
Each VAL question is reworded with another template frame for its type, keeping its image and label; for counting, the false number asked is also redrawn, so that true and false numbers share the range 1--7, and the singular/plural contrast is removed.
The design and scoring were registered before any VLM was evaluated on it, with one amendment, which set the range of the false counts.
It re-encodes rather than removes the counting cue: answering \yes exactly when the asked number is one solves \vf{78.7\%} of the counting pairs in both versions, and the frozen text-only spatial classifier still answers \vf{80.3\%} of the spatial pairs, so the rephrased set is not cue-free.

\emph{Rewritten gold spatial questions.}
Each of the \vf{150} gold spatial pairs is rewritten twice from its stored slots: with a paraphrase template from the rephrased VAL above, and as the converse relation, which swaps the two objects and inverts the relation; the gold labels carry over unchanged.
The author reviewed the text of all \vf{600} rewritten questions and answered \vf{50} converse pairs blind from the images; these answers agree with the carried-over labels on \vf{90} of \vf{100} questions and \vf{41} of \vf{50} pairs.
The design and two findings were registered before the rewrites were generated, and the rewrites were frozen before any model saw them: the gain survives a version if \ours's difference from the base model on it has an interval above zero (it does on both: \vf{87.3\%} \vs\ \vf{62.0\%} and \vf{94.5\%} \vs\ \vf{70.0\%}), and it is tied to our templates if \ours's change from the original wording has an interval below zero (it is, for the paraphrase).

\emph{What'sUp.}
Its two controlled subsets have \vf{820} photos of \vf{205} object pairs, one photo per relation, and none matches an image of ours by source identity, SHA-256, or perceptual hash.
In the yes/no format, each photo gets two questions in our spatial template, one true and one false, and counts as right only if both answers are; in caption choice, the four captions of its subset are offered in four rotated orders, and accuracy is averaged over the orders.
Every caption-choice response is free text, so all of them go to the locked extractor.
The test, both formats, and the breakdowns by relation were registered after the main results and before any model was evaluated on What'sUp; intervals use \vf{2{,}000} bootstrap draws over the \vf{205} object pairs (seed 20261109).

\emph{Held-out benchmarks.}
Their tiers, scorers, populations, and bootstrap procedure were registered before any adapter was evaluated on them, with contrasts against \ours; the other arms' contrasts with the base model and the three-seed means were added later.
The official tier uses each benchmark's own parser; the robust tier scores a response that is exactly one option label directly and sends any other response to the extractor; unresolved responses count as wrong in both.
BLINK is the authors' macro over its tasks, CV-Bench 2D the mean of its COCO and ADE20K parts, and MME a score, not a percentage.
Items that share an image with any of our training, replay, selection, or development data, by canonical identity, SHA-256, or perceptual hash, are left out together with every item linked to them through a shared image: MME keeps \vf{2{,}310} of its \vf{2{,}374} items, CV-Bench 2D \vf{1{,}260} of \vf{1{,}438}, BLINK \vf{1{,}772} of \vf{1{,}901}, and CV-Bench 3D and MMVP all of theirs.
Contrasts are paired bootstrap intervals over shared image clusters (\vf{2{,}000} replicates, seed 20261014).

\emph{Answer extraction.}
The extractor is Muse Glimmer~30B~\cite{muse} at a pinned revision (\texttt{a4e59da5}), run in bf16 with greedy decoding, batch size one, and seed 42.
The author checked it twice against a reading of the same responses: it chose the same option for \vf{79} of \vf{80} base-model responses to capability probes (\vf{50} VisMin and \vf{30} NegVQA), and for \vf{59} of \vf{60} free-form base-model responses to the held-out benchmarks, half of which the official parsers could not resolve (registered threshold: \vf{57}).
These checks test whether the extractor reads a response faithfully, not whether the response is right.

\section{Additional results}
\label{supp:results}

\emph{Repair by type.}
\cref{tab:valtypes} splits \ours's VAL repair by type (five seeds): spatial relations are repaired most and depend most on the image (a gap of \vf{64.2} points, from \vf{48.7} to \vf{77.0} across seeds), whereas counting and negation keep most of their accuracy with a black image.
\begin{table}[t]
  \caption{\textbf{VAL repair by type.} Pair accuracy of \ours on VAL with the image and with a black one (\%, mean of five seeds; 300 pairs per type) and their difference in points.}
  \label{tab:valtypes}
  \centering
  \footnotesize
  \setlength{\tabcolsep}{4pt}
  \begin{tabular}{@{}l ccc@{}}
    \toprule
    Type & Repair & Black image & Gap \\
    \midrule
    Object existence & \vf{76.3} & \vf{30.3} & \vf{46.0} \\
    Attribute        & \vf{66.1} & \vf{38.3} & \vf{27.7} \\
    Spatial relation & \vf{93.0} & \vf{28.8} & \vf{64.2} \\
    Counting         & \vf{90.0} & \vf{76.6} & \vf{13.4} \\
    Negation         & \vf{88.9} & \vf{63.7} & \vf{25.2} \\
    \midrule
    All & \vf{82.8} & \vf{47.5} & \vf{35.3} \\
    \bottomrule
  \end{tabular}
\end{table}

\emph{AMBER and NegVQA.}
On AMBER, scored at one decimal, \ours (seed 42) moves CHAIR from \vf{8.2} to \vf{8.0} (lower is better), coverage from \vf{72.3} to \vf{72.8}, the hallucination rate from \vf{58.6} to \vf{57.0}, and discriminative accuracy from \vf{88.3} to \vf{88.9}.
On NegVQA's negated questions (seed 42), \ours improves from \vf{62.3\%} to \vf{69.6\%} (\uv: \vf{69.3\%}; \oursmi: \vf{68.3\%}), but about \vf{72\%} of its gain comes from committing to an answer more often rather than from more accurate answers.

\emph{What'sUp by relation.}
\cref{tab:whatsup} splits What'sUp's yes/no results by relation (a registered breakdown; intervals over object pairs).
Left/right and in front of/behind stay at ceiling for every model.
The arms trained on all five of our types keep or raise their accuracy on photos of an object on another (\ours: \vf{91.1\%}; base model: \vf{86.4\%}) but fail on photos of it under the other: random pairs least (\vf{45.6\%}; \ours, seed 42: \vf{19.4\%}, interval for the difference \vf{16.8}--\vf{36.6} points), \ours (five seeds) \vf{26.4\%}, and \uv and \ours without replay most (\vf{2.3\%} and \vf{7.8\%}); multi-image replay does not help (\vf{17.5\%}, seed 42).
\vbc, trained on VisMin's edits in our format, loses on both (\vf{63.8\%} and \vf{71.8\%}); replay alone and \vbn gain on ``under'' (\vf{88.3\%} and \vf{87.7\%}), and their changes on ``on'' (\vf{87.4\%} and \vf{82.5\%}) have intervals that include zero.
With caption choice, the only changes for \ours (five seeds) whose intervals exclude zero are a loss on left/right (\vf{99.5\%} to \vf{99.0\%}; interval for the change \vf{$-0.9$} to \vf{$-0.2$}) and a gain on ``behind'' (\vf{91.4\%} to \vf{94.7\%}; \vf{$+1.6$} to \vf{$+5.1$}); it picks the right caption for every ``on'' and ``under'' photo (base model: \vf{99.8\%} and \vf{100\%}), and the share of photos it gets right in all four option orders is unchanged (\vf{94.0\%}).
The seed-42 controls show the same split: random pairs and \uv, which answer both yes/no questions correctly for only \vf{45.6\%} and \vf{1.9\%} of the ``under'' photos, pick the right caption for every one of them, as does \vbc (yes/no: \vf{74.8\%}), and replay alone misses one option order of one photo (\vf{99.8\%}); \uv also loses slightly on left/right (\vf{98.9\%} \vs\ \vf{99.5\%}; interval \vf{$-1.0$} to \vf{$-0.2$}), while random pairs and replay alone show no clear change there.
So do the remaining arms: \oursmi (seed 42) and \uv's other two seeds pick the right caption for every ``under'' photo, training without replay misses \vf{2} of the \vf{412} option orders (yes/no: \vf{7.8\%}), and the VisMin arms miss none; pooled, the three-seed means stay near the base model's \vf{97.6\%} (\uv: \vf{97.8\%}; \vbn: \vf{97.9\%}; \vbc: \vf{97.6\%}; intervals include zero), and single seeds range from \vf{97.1\%} to \vf{98.4\%}.
\begin{table}[t]
  \caption{\textbf{What'sUp in the yes/no format, by relation}: the share of photos whose two questions, about the true relation and its opposite, are both answered correctly; all \vf{820} photos, left/right (\vf{410}), on (\vf{103}), and under (\vf{103}).
  In front of and behind: \vf{98.0}--\vf{100\%} for every model.
  Seeds as in \cref{tab:main}; no replay: seed 42.}
  \label{tab:whatsup}
  \centering
  \footnotesize
  \setlength{\tabcolsep}{4.0pt}
  \begin{tabular}{@{}l cccc@{}}
    \toprule
    Model & All & Left/right & On & Under \\
    \midrule
    \base                & \vf{95.6} & \vf{99.0} & \vf{86.4} & \vf{84.5} \\
    Replay only          & \vf{96.2} & \vf{99.0} & \vf{87.4} & \vf{88.3} \\
    \vbn                 & \vf{95.7} & \vf{99.0} & \vf{82.5} & \vf{87.7} \\
    \vbc                 & \vf{91.4} & \vf{99.0} & \vf{63.8} & \vf{71.8} \\
    \uv                  & \vf{87.1} & \vf{99.5} & \vf{96.8} & \vf{2.3} \\
    Random pairs         & \vf{92.1} & \vf{99.5} & \vf{93.2} & \vf{45.6} \\
    No replay            & \vf{87.9} & \vf{99.5} & \vf{98.1} & \vf{7.8} \\
    \midrule
    \ours                & \vf{89.4} & \vf{99.5} & \vf{91.1} & \vf{26.4} \\
    \oursmi, seed 42     & \vf{88.3} & \vf{99.5} & \vf{91.3} & \vf{17.5} \\
    \bottomrule
  \end{tabular}
\end{table}

\begin{table*}[t]
  \caption{\textbf{What'sUp in the yes/no format, by question}: the share of photos (\%) for which each question is answered yes, for the \vf{103} ``on'' and the \vf{103} ``under'' photos, and the share of the \vf{206} ``on'' and the \vf{206} ``under'' questions answered yes with a black image (post hoc).
  On each kind of photo, the first two questions use What'sUp's relation words and the last two the words of our relation pairs, with What'sUp's relations as labels; of each two, the first is true and the second false, and with a black image every question is false.
  A tie between the two scores is not a yes.
  Seeds as in \cref{tab:main}; no replay and no directional pairs: seed 42.}
  \label{tab:onunder}
  \centering
  \footnotesize
  \setlength{\tabcolsep}{4.5pt}
  \begin{tabular}{@{}l cccc cccc cc@{}}
    \toprule
    & \multicolumn{4}{c}{``On'' photos} & \multicolumn{4}{c}{``Under'' photos} & \multicolumn{2}{c}{Black image} \\
    \cmidrule(lr){2-5}\cmidrule(lr){6-9}\cmidrule(l){10-11}
    Question about & on & under & above & below & under & on & below & above & on & under \\
    \midrule
    \base                & \vf{99.0} & \vf{12.6} & \vf{99.0} & \vf{34.0} & \vf{96.1} & \vf{11.7} & \vf{98.1} & \vf{1.9} & \vf{0} & \vf{0} \\
    Replay only          & \vf{99.0} & \vf{11.7} & \vf{99.0} & \vf{25.2} & \vf{96.1} & \vf{7.8}  & \vf{97.1} & \vf{1.0} & \vf{0} & \vf{0} \\
    \vbn                 & \vf{99.0} & \vf{16.2} & \vf{87.1} & \vf{23.0} & \vf{97.1} & \vf{9.1}  & \vf{97.7} & \vf{0}   & \vf{0} & \vf{0} \\
    \vbc                 & \vf{99.0} & \vf{31.7} & \vf{70.9} & \vf{10.4} & \vf{96.8} & \vf{22.3} & \vf{97.1} & \vf{0}   & \vf{0} & \vf{0} \\
    \uv                  & \vf{100}  & \vf{2.9}  & \vf{44.7} & \vf{12.0} & \vf{100}  & \vf{97.7} & \vf{100}  & \vf{0}   & \vf{71.8} & \vf{69.9} \\
    Random pairs         & \vf{99.0} & \vf{5.8}  & \vf{57.3} & \vf{1.9}  & \vf{98.1} & \vf{50.5} & \vf{100}  & \vf{0}   & \vf{0} & \vf{0} \\
    No replay            & \vf{100}  & \vf{1.9}  & \vf{65.0} & \vf{0}    & \vf{100}  & \vf{90.3} & \vf{100}  & \vf{0}   & \vf{71.8} & \vf{56.3} \\
    No directional pairs & \vf{99.0} & \vf{86.4} & \vf{99.0} & \vf{95.1} & \vf{98.1} & \vf{79.6} & \vf{98.1} & \vf{21.4} & \vf{29.1} & \vf{9.7} \\
    \midrule
    \ours                & \vf{100}  & \vf{8.7}  & \vf{46.0} & \vf{17.9} & \vf{99.8} & \vf{70.3} & \vf{99.6} & \vf{0}   & \vf{32.2} & \vf{32.8} \\
    \oursmi, seed 42     & \vf{100}  & \vf{8.7}  & \vf{50.5} & \vf{0}    & \vf{100}  & \vf{79.6} & \vf{99.0} & \vf{0}   & \vf{36.9} & \vf{31.1} \\
    \bottomrule
  \end{tabular}
\end{table*}

\emph{Which question fails.}
\cref{tab:onunder} gives, for the ``on'' and ``under'' photos, how often each of a photo's two questions is answered yes, first in a post hoc breakdown of the same predictions and then in the two conditions described below.
The photo of a scarf under a table, for example, gets ``In the image, is there a scarf under a table?'' (\yes) and ``In the image, is there a scarf on a table?'' (\no).
In these predictions, every model answers yes to nearly every true question (at least \vf{96.1\%}).
The arms trained on all five of our types fail on one of them: on ``under'' photos they also answer yes to the false ``on'' question, \ours for \vf{70.3\%} of the photos (base model: \vf{11.7\%}; difference \vf{58.6} points, interval \vf{50.4} to \vf{67.0}), \uv for \vf{97.7\%}, training without replay for \vf{90.3\%}, \oursmi for \vf{79.6\%}, and random pairs for \vf{50.5\%} (for each, the interval for the difference from the base model starts at \vf{29.9} points or more).
On ``on'' photos, none of these arms, averaged over its seeds, answers yes to the false ``under'' question more often than the base model does (\vf{12.6\%}; \ours: \vf{8.7\%}, an interval including zero).
\vbc accepts both false questions more often than the base model (\vf{31.7\%} and \vf{22.3\%}; intervals above zero), and replay alone and \vbn accept the false ``on'' question no more often (\vf{7.8\%} and \vf{9.1\%}).
\ours's yes to the false ``on'' question is also close to the decision boundary: its mean log-probability margin of yes over no on that question is \vf{$+1.3$} (base model: \vf{$-7.5$}), against \vf{$+7.7$} on the true ``under'' question, and for a further \vf{3.1\%} of the photos the two scores tie, which counts as wrong.
None of the \vf{24{,}000} repair questions that \ours trains on contains ``under'', ``underneath'', ``beneath'', or ``on top of'' (``above'': \vf{1{,}857}; ``below'': \vf{1{,}815}); of its \vf{12{,}000} replay rows, \vf{53} contain ``under'', \vf{9} of them with a yes/no target (\vf{5} yes).

\emph{Other words and no image.}
We then asked each of these photos' two questions with the words of our relation pairs, ``above'' for ``on'' and ``below'' for ``under'', keeping What'sUp's relations as labels without re-annotating them, and asked the original questions with a black image, on which every one of them is false (post hoc; specified before the runs).
With ``below'' and ``above'', the base model and every arm of \cref{tab:whatsup} agree with both carried-over labels for nearly every ``under'' photo (at least \vf{96.1\%}; \ours: \vf{99.6\%}; the run without directional relation pairs described below: \vf{75.7\%}).
On the ``on'' photos, where the carried-over ``below'' label can be ambiguous (a chair's back, for example, rises above an object on its seat), the base model affirms ``below'' for \vf{34.0\%}; averaged over their seeds, the arms trained on all five of our types affirm it less often but deny ``above'' far more often: \ours affirms it for \vf{46.0\%} of these photos (base model: \vf{99.0\%}; difference \vf{$-53.0$} points, interval \vf{$-61.2$} to \vf{$-44.9$}) and agrees with both labels for \vf{42.5\%} (base model: \vf{62.1\%}).
With a black image, the base model answers no to all \vf{412} questions, as do replay alone, random pairs, and the VisMin arms; \ours answers yes to \vf{32.2\%} of the ``on'' and \vf{32.8\%} of the ``under'' questions (seeds: \vf{15.5}--\vf{41.7\%} and \vf{10.7}--\vf{66.0\%}), with mean margins near zero (\vf{$-0.5$} for both; base model: \vf{$-13.8$} and \vf{$-11.3$}), and \uv and training without replay answer yes to more than half.
So the false ``on'' answers depend on the photo, not on the question alone: random pairs answer no to every question on a black image yet affirm the false ``on'' for half of the ``under'' photos, and \ours affirms it for \vf{70.3\%} of those photos against \vf{32.2\%} of all ``on'' questions on a black image.

\emph{No directional relation pairs.}
To see whether the error needs our directional relation pairs, we trained one more arm on \ours's recipe and seed 42 without them (post hoc; specified before training): its repair data drop all \vf{2{,}400} spatial pairs and the \vf{1{,}021} negation pairs whose questions use ``to the left of'', ``to the right of'', ``above'', or ``below'', keeping \vf{20} negation pairs about overlapping objects, and its \vf{8{,}579} remaining pairs fill the same \vf{12{,}000} presentations (\vf{3{,}000} per type, \vf{1.4} per pair on average) next to the same replay rows, which include a few spatial questions.
It shows the error and its mirror image: it affirms the false ``on'' for \vf{79.6\%} of the ``under'' photos, as often as \ours with the same seed (\vf{78.6\%}; difference \vf{$+1.0$} points, interval \vf{$-2.1$} to \vf{$+4.5$}), and the false ``under'' for \vf{86.4\%} of the ``on'' photos (\ours, seed 42: \vf{6.8\%}; difference \vf{79.6} points, interval \vf{71.4} to \vf{87.0}), so it answers both questions correctly for only \vf{17.5\%} and \vf{8.7\%} of them.
With ``above'' and ``below'', it affirms ``below'', labeled false, for \vf{95.1\%} of the ``on'' photos; only ``above'' on ``under'' photos is mostly rejected (affirmed for \vf{21.4\%}).
Its What'sUp yes/no accuracy falls to \vf{76.8\%} (base model: \vf{95.6\%}), while left/right stays at \vf{99.0\%}.
In this run, training on our other four types thus led the model to affirm vertical relations in these photos largely whichever is true, while the arms that also train on directional relation pairs usually reject the questions labeled false for ``under'', ``below'', and ``above'' but often affirm the false ``on''.
Because the run also changes the mix of types and repeats pairs, and has one seed, we do not attribute its difference from \ours to the directional relation pairs alone.
Its VAL pair accuracy is \vf{67.8\%} (\ours, five seeds: \vf{82.8\%}), and it repairs \vf{38.7\%} of the spatial VAL pairs without directional relation pairs (\vf{9.0\%} with a black image; with the image, base model: \vf{7.7\%}; \ours: \vf{93.0\%}).

\section{VisMin baselines}
\label{supp:vismin}

\emph{Data.}
We decoded and hashed all \vf{110{,}497} images of VisMin's training release; of its \vf{64{,}392} edits, \vf{41{,}117} in \vf{30{,}218} source-edit families remain after we drop every family whose images occur in our VAL or TEST, the replay release, or the benchmarks (by lineage, SHA-256, and perceptual hash).
VisMin data is never filtered with our committee.
\vbc takes \vf{3{,}000} edits each of objects, attributes, counting, and relations and turns each into a minimal pair in our format, asked about the source image for half of them and the edited image for the other half (\vf{12{,}000} pairs).
\vbn takes \vf{1{,}500} of these edits per category, each giving one image-choice and one caption-choice item in the authors' format, rebuilt with their pinned builder; we omit its captioning rows (\vf{12{,}000} items).
Both train with \ours's recipe and budget: \vf{1{,}500} updates of 8 repair items and the same 8 replay rows.

\emph{Conversion.}
\vbc asks \texttt{Does the image match this description: "\{caption\}"?} with our training wrapper, and its targets are \texttt{Yes, the image matches this description: "\{true caption\}".} and \texttt{No, the image matches this description: "\{true caption\}".}, with the same decision weighting as ours; no model rewrites a caption.
\vbn keeps the authors' choice questions and answers and their uniform token loss.

\emph{Tuning.}
The reported VisMin arms use \ours's learning rate and batch size.
A registered learning-rate and batch-size search for \vbn found nothing to adopt: four settings with larger batches (16 or 32 repair units with as many replay rows, at learning rates $10^{-4}$ and $2{\times}10^{-4}$) were excluded under an amended rule, registered before they ran, that allows at most \vf{25} answers hitting the generation cap in any \vf{100} consecutive answers (they had \vf{40} to \vf{66} in their first \vf{100}), and the best eligible setting, learning rate $10^{-4}$ at our batch size, gained \vf{0.76} points on the registered development objective, below the \vf{0.84}-point threshold for adoption review.

\section{Two-image side-effect analyses}
\label{supp:sideeffect}

\emph{Decomposition.}
Paired accuracy is the share of groups in which a model picks different images for the two opposite captions (caption sensitivity) times its accuracy within those groups (\cref{tab:sideeffect}): \vf{77.2\%} and \vf{96.3\%} for the base model, \vf{48.6\%} and \vf{97.0\%} for \ours (seed 42).
Training removes caption sensitivity, not accuracy where the model is still sensitive.

\emph{Other arms.}
Training without replay falls further (\vf{41.5\%} paired; caption sensitivity \vf{42.9\%}) although its answers are longer (median \vf{69} tokens), and \uv's seed 44 falls like its seed 42 (\vf{49.2\%}; \vf{2} tokens).
VisMin's relation test is in-distribution for the VisMin arms (seed 42): \vbc, trained on VisMin's edits in our single-image format, scores \vf{52.4\%} (\vf{2} tokens), and \vbn, trained in VisMin's own format, \vf{92.1\%} (\vf{67} tokens).
For all four, accuracy within caption-sensitive groups stays at \vf{96.2}--\vf{97.4\%}.

\emph{Relation axes.}
VisMin's \vf{622} relation groups compare left and right (\vf{380}), above and below (\vf{225}), or other relations (\vf{17}).
The base model is right on \vf{83.2\%} of the left/right and \vf{61.8\%} of the above/below groups; \ours ends near half on both (five seeds: \vf{50.3\%} each, every seed within \vf{46.7}--\vf{54.2\%}), as does replay alone (\vf{50.5\%} and \vf{47.1\%}), while \oursmi's seed 42 keeps \vf{82.9\%} and \vf{66.7\%}.
Of the \vf{169} groups that \ours (seed 42) loses on net (\vf{209} lost, \vf{40} gained), \vf{133} are left/right (\vf{143} lost, \vf{10} gained) and \vf{34} above/below (\vf{61} lost, \vf{27} gained).

\emph{Probes.}
Two diagnostic probes of \ours (seed 42), not native VisMin scores, test what changed.
Asking it to describe each image first raises caption sensitivity by \vf{16.1} points and paired accuracy by \vf{15.4}, and none of its \vf{1{,}244} answers then stays at four tokens or fewer.
Swapping the order of the two images shows whether its choices follow the position or the image: of the \vf{1{,}241} requests it answers in both orders, \vf{818} keep the same image and \vf{423} the same position, so it still mostly tracks the images, although it picks the first image more often than the base model (\vf{60.5\%} against \vf{51.4\%} of answers).
A human check of \vf{40} answers of the trained models on this task (\ours and \uv, seed 42; every free-form answer and seeded short ones) agreed with the extractor on all \vf{40}, so the drop is not an extraction artifact.

\emph{Multi-image replay data.}
\oursmi replaces \vf{1{,}500} of the description rows in \ours's replay data, chosen by a seeded hash, with two-image caption choices built from unedited COCO photographs and their human captions: \vf{750} pairs of photos that share their largest object category and overlap in their object sets, each asked once for each photo's caption.
The base model's own answers are the targets (greedy, at most \vf{512} new tokens), kept only if both answers of a pair choose the right photo; \vf{1{,}425} of \vf{1{,}500} candidate pairs passed, and the first \vf{750} were used.
The other \vf{10{,}500} replay rows, the repair data, and the objective are unchanged.

\fi

\end{document}